%% file: arxiv.tex
\documentclass[10pt,twocolumn,letterpaper]{article}

\usepackage[accsupp]{axessibility}  %

\usepackage{wacv}
\usepackage{times}
\usepackage{epsfig}
\usepackage{graphicx}
\usepackage{amsmath}
\usepackage{amssymb}
\usepackage{booktabs}
\usepackage{subfig}
\usepackage[normalem]{ulem}
\usepackage{tikz}
\usepackage{comment}

\usepackage{enumitem} %
\usepackage[linesnumbered,ruled,vlined]{algorithm2e} %

\wacvfinalcopy %

\ifwacvfinal
\usepackage[breaklinks=true,bookmarks=false]{hyperref}
\else
\usepackage[pagebackref=true,breaklinks=true,colorlinks,bookmarks=false]{hyperref}
\fi

\begin{document}

\title{Scaling Neural Face Synthesis to High FPS and Low Latency by Neural Caching}

\author{
Frank Yu\\
\\
{\tt\small frankyu@cs.ubc.ca}
\and
Sid Fels\\
University of British Columbia (UBC)\\  
{\tt\small ssfels@ece.ubc.ca}\\ 
\\
\and
Helge Rhodin\\
\\
{\tt\small rhodin@cs.ubc.ca}
}

\input{macros}

\maketitle
\thispagestyle{empty}

\input{00_abstract}
\input{01_introduction}
\input{02_related}
\input{03_method}

\input{04_experiments}

\input{05_results}

\input{06_discussion}
\input{07_conclusion}

\input{08_supplemental_arxiv}

{\small
\bibliographystyle{ieee_fullname}
\bibliography{egbib}
}

\end{document}

%% file: macros.tex
\newcommand{\red}[1]{{\color{red}#1}}
\newcommand{\tb}[1]{\textbf{#1}}
\newcommand{\tbi}[1]{\textit{\textbf{#1}}}
\newcommand{\norm}[1]{\left\|#1\right\|}

\newcommand{\figref}[1]{Figure~\ref{#1}}
\newcommand{\secref}[1]{Section~\ref{#1}}
\newcommand{\feqref}[1]{Equation~\eqref{#1}}
\newcommand{\tabref}[1]{Table~\ref{#1}}

\newcommand{\VOOne}[1]{{1x warp}}
\newcommand{\VTTwo}[1]{{2x warp}}

\newcommand{\rev}[1]{{\color{red}#1}}

\newcommand{\hmvec}[1]{
\begin{bmatrix}#1 \\ 1
\end{bmatrix}
}

\newcommand{\parag}[1]{\textbf{#1}}%

\newcommand{\veclist}[2]{
\left[#1_1,#1_2,\cdots,#1_{#2}\right]
}
\newcommand{\veclistki}[2]{
[#1_{k,1},\cdots,#1_{k,#2}]
}
\newcommand{\veclistkiz}[2]{
[#1_{k,0},\cdots,#1_{k,#2}]
}
\newcommand{\subki}[1]{#1_{k,i}}
\newcommand{\subkj}[1]{#1_{k,j}}
\newcommand{\subkm}[1]{#1_{k,m}}
\newcommand{\subkl}[1]{#1_{k,l}}

\newcommand{\wpe}[2]{\gamma\left(#1_{k,#2},w_{k,#2}\right)}
\newcommand{\pesin}[2]{\sin\left(2^{#1}\pi#2\right)}
\newcommand{\pecos}[2]{\cos\left(2^{#1}\pi#2\right)}

\definecolor{olive}{RGB}{50,150,50}
\definecolor{frank}{RGB}{198, 3, 252}

\newcommand{\TODO}[1]{\textcolor{red}{TODO: #1}}
\newcommand{\todo}[1]{\textcolor{red}{#1}}

\newcommand{\R}{\mathbb{R}}
\newcommand{\argmin}{\operatornamewithlimits{argmin}}

\newcommand{\supp}{supplemental material}

\newcommand{\mA}{\mathbf{A}}
\newcommand{\mB}{\mathbf{B}}
\newcommand{\mC}{\mathbf{C}}
\newcommand{\mD}{\mathbf{D}}
\newcommand{\mE}{\mathbf{E}}
\newcommand{\mF}{\mathbf{F}}
\newcommand{\mG}{\mathbf{G}}
\newcommand{\mGamma}{\mathbf{\Gamma}}
\newcommand{\mH}{\mathbf{H}}
\newcommand{\mI}{\mathbf{I}}
\newcommand{\mJ}{\mathbf{J}}
\newcommand{\mK}{\mathbf{K}}
\newcommand{\mL}{\mathbf{L}}
\newcommand{\mM}{\mathbf{M}}
\newcommand{\mN}{\mathbf{N}}
\newcommand{\mO}{\mathbf{O}}
\newcommand{\mP}{\mathbf{P}}
\newcommand{\mQ}{\mathbf{Q}}
\newcommand{\mR}{\mathbf{R}}
\newcommand{\mRvr}{\mathbf{R}_{\text{virt}\rightarrow\text{real}}}
\newcommand{\mRrv}{\mRvr^{-1}}%
\newcommand{\mS}{\mathbf{S}}
\newcommand{\mT}{\mathbf{T}}
\newcommand{\mU}{\mathbf{U}}
\newcommand{\mV}{\mathbf{V}}
\newcommand{\mW}{\mathbf{W}}
\newcommand{\mX}{\mathbf{X}}
\newcommand{\mY}{\mathbf{Y}}
\newcommand{\mZ}{\mathbf{Z}}

\newcommand{\cA}{\mathcal A}
\newcommand{\cB}{\mathcal B}
\newcommand{\cC}{\mathcal C}
\newcommand{\cD}{\mathcal D}
\newcommand{\cE}{\mathcal E}
\newcommand{\cF}{\mathcal F}
\newcommand{\cG}{\mathcal G}
\newcommand{\cH}{\mathcal H}
\newcommand{\cI}{\mathcal I}
\newcommand{\cJ}{\mathcal J}
\newcommand{\cK}{\mathcal K}
\newcommand{\cL}{\mathcal L}
\newcommand{\cM}{\mathcal M}
\newcommand{\cN}{\mathcal N}
\newcommand{\cO}{\mathcal O}
\newcommand{\cP}{\mathcal P}
\newcommand{\cQ}{\mathcal Q}
\newcommand{\cR}{\mathcal R}
\newcommand{\cS}{\mathcal S}
\newcommand{\cT}{\mathcal T}
\newcommand{\cU}{\mathcal U}
\newcommand{\cV}{\mathcal V}
\newcommand{\cW}{\mathcal W}
\newcommand{\cX}{\mathcal X}
\newcommand{\cY}{\mathcal Y}
\newcommand{\cZ}{\mathcal Z}

\newcommand{\va}{\mathbf{a}}
\newcommand{\vb}{\mathbf{b}}
\newcommand{\vc}{\mathbf{c}}
\newcommand{\vd}{\mathbf{d}}
\newcommand{\ve}{\mathbf{e}}
\newcommand{\vf}{\mathbf{f}}
\newcommand{\vg}{\mathbf{g}}
\newcommand{\vh}{\mathbf{h}}
\newcommand{\vi}{\mathbf{i}}
\newcommand{\vj}{\mathbf{j}}
\newcommand{\vk}{\mathbf{k}}
\newcommand{\vl}{\mathbf{l}}
\newcommand{\vm}{\mathbf{m}}
\newcommand{\vn}{\mathbf{n}}
\newcommand{\vo}{\mathbf{o}}
\newcommand{\vp}{\mathbf{p}}
\newcommand{\vq}{\mathbf{q}}
\newcommand{\vr}{\mathbf{r}}
\renewcommand{\vs}{\mathbf{s}}
\newcommand{\vt}{\mathbf{t}}
\newcommand{\vu}{\mathbf{u}}
\newcommand{\vv}{\mathbf{v}}
\newcommand{\vw}{\mathbf{w}}
\newcommand{\vx}{\mathbf{x}}
\newcommand{\vy}{\mathbf{y}}
\newcommand{\vz}{\mathbf{z}}

\newcommand{\bR}{\mathbb{R}}
\newcommand{\mx}{\mathbf{x}}
\newcommand{\mj}{\mathbf{j}}

\newcommand{\RNum}[1]{\uppercase\expandafter{\romannumeral #1\relax}}

%% file: 00_abstract.tex
\begin{abstract}
Recent neural rendering approaches greatly improve image quality, reaching near photorealism. However, the underlying neural networks have high runtime, precluding telepresence and virtual reality applications that require high resolution at low latency. The sequential dependency of layers in deep networks makes their optimization difficult. We break this dependency by caching information from the previous frame to speed up the processing of the current one with an implicit warp. The warping with a shallow network reduces latency and the caching operations can further be parallelized to improve the frame rate. In contrast to existing temporal neural networks, ours is tailored for the task of rendering novel views of faces by conditioning on the change of the underlying surface mesh. We test the approach on view-dependent rendering of 3D portrait avatars, as needed for telepresence, on established benchmark sequences. Warping reduces latency by 70$\%$ (from 49.4ms to 14.9ms on commodity GPUs) and scales frame rates accordingly over multiple GPUs while reducing image quality by only 1$\%$, making it suitable as part of end-to-end view-dependent 3D teleconferencing applications. Our project page can be found at: \color{blue}{\href{https://yu-frank.github.io/lowlatency/}{https://yu-frank.github.io/lowlatency/}}.

\end{abstract}

%% file: 01_introduction.tex
\section{Introduction}

\begin{figure}[t!]
\begin{center}
\includegraphics[width=0.95\linewidth, trim={1cm 39.25cm 7.55cm 0cm},clip]{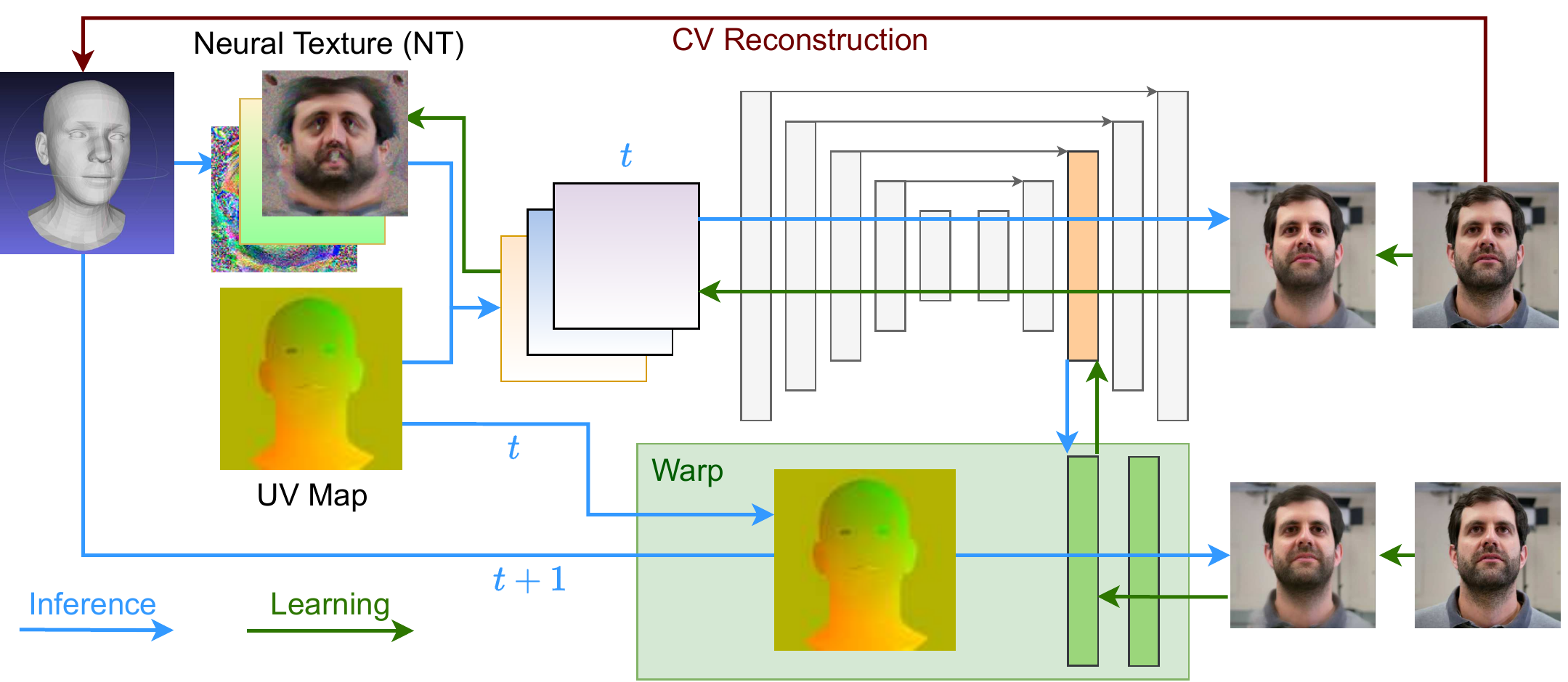}
\end{center}
\caption{\label{fig:neural rendering}
		{\bf Neural Rendering with Warping.} While a single frame method requires a deep neural network to synthesize a realistic head image from rough geometry, our implicit warping yields low latency for a new frame at $t+1$ with a shallow network.
	}
\end{figure}

Telepresence via photo-realistic 3D avatars promises to better connect people. Recent advances in neural rendering already enable near photo-realistic image quality, but the underlying deep neural networks limit the best possible latency with their sequential, layer-wise processing.
This is a problem for virtual reality applications as these require low latency upon head motion of the user to convey an immersive experience. For example, \cite{zhou2019evaluation} creates a high-fidelity system using 120Hz projectors and user viewpoint tracking with a tracker having 60Hz updates and 3ms latency to minimize users' perception of warping of the scene when they move. 
However, none of the existing neural renderers reaches the required \emph{motion to photons latency}, i.e., the time it takes from the user input, such as moving the head in VR, until the display updates.
It is henceforth an open problem to find improved neural models that strike a better trade-off between speed and image quality.

We develop a parallel implementation, inspired by previous approaches for improving the frame rate for other video processing tasks \cite{carreira2018massively} by spreading the computation over several frames over multiple GPUs. By itself, parallel execution only improves frame rate, not the latency that is so crucial for VR, since the effective network depth of sequentially executed layers remains the same. 
To nevertheless reach the desired latency reduction and frame rate, we combine parallel execution with a dedicated warp layer that is tailored for neural face synthesis and acts as a skip connection between consecutive frames.

This neural caching approach re-uses information from the previous time step to improve latency from the current one with a shallow network for image generation and a deep network for computing the cache while waiting for the next input frame. Note that this caching strategy by itself already improves latency on a single GPU and scales in frame rate on increasingly parallel hardware by breaking the sequential layer dependencies and offloading the image cache generation into multiple threads.

The difficulty lies in finding the right representation for the cache to succeed with a low-capacity network. 
Our approach is inspired by classical low-latency rendering on VR displays~\cite{van2016asynchronous}, which compute only the position of objects for in-between frames and warp color from the previous. However, explicit warping does not apply well to neural renderers where the underlying geometry is approximate and the neural texture is high-dimensional making warping operations much more costly.

To warp the neural representation, we introduce an implicit warp that provides a skip connection that is tailored for neural rendering by taking into account the head model parameterization.
The result is a warp network that models the image-space motion from one frame to the next given the desired viewpoint and head model parameters. Figure~\ref{fig:neural rendering} outlines the main components.

\begin{figure}[t!]
\begin{center}
\includegraphics[height=0.45\linewidth, trim={0cm 0.5cm 0cm 0cm},clip]{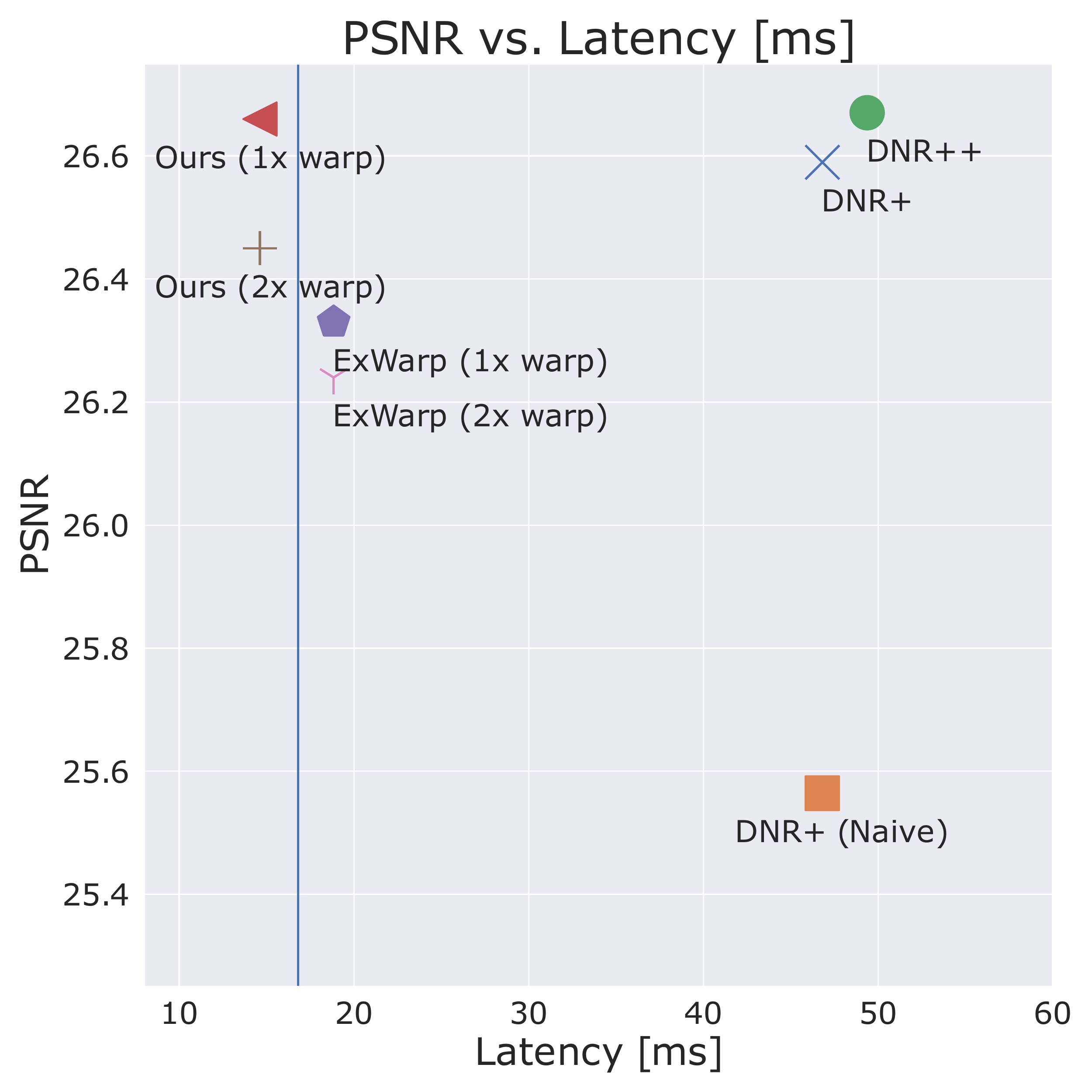}
\includegraphics[height=0.45\linewidth, trim={0cm 0.5cm 0cm 0cm},clip]{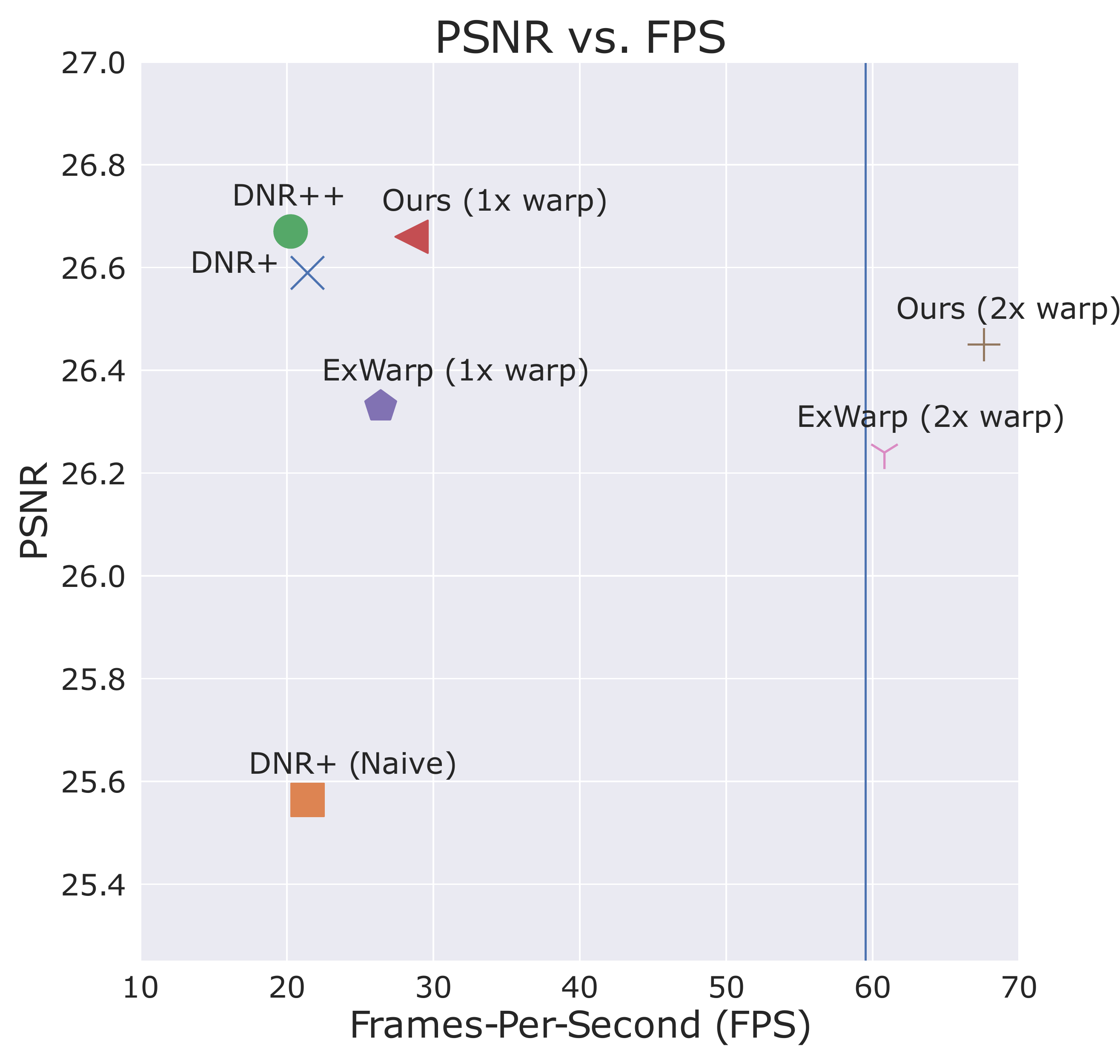}
\end{center}
\caption{\label{fig:pref-v-lat}
		{\bf Performance vs. latency (left) and fps (right)} for our models and the baselines on the \emph{beard} dataset. Optimal performance is in the top-left and top-right corners, indicated by the yellow star in each respective figure. The vertical blue line marks the latency and fps of \cite{zakharov2020fast}. Models suitable for parallel execution are run using two GPUs.}
\end{figure}

Our design is geared towards \emph{novel-view-synthesis} of a talking head given a dynamically changing viewpoint, such as the user's head motion in VR. We build upon the deferred neural renderer (DNR)~\cite{thies2019deferred} that uses a neural texture learned at training time. Our contributions towards scaling frame rate and latency on parallel hardware are:
\begin{itemize}[noitemsep]
\item Demonstrating that the proposed neural caching can reduce latency by up to 70\% with minimal degradation in image quality (only 1\% PSNR).
    \item Extending a DNR to generate a high-resolution output with low latency via caching and an implicit warping.
    \item Developing a parallel scheduler that supports warping and synchronizes multiple threads using queues.
    \item Refining the representation of facial expression, head pose, and camera angle to improve the implicit warp.
    \item Adding head-stabilization and tweaking the backbone and training strategy for noisy real-world conditions.
\end{itemize}%
Our experiments highlight the importance of how to cache as well as what and how information from the current frame is passed to the shallow \emph{implicit warping network}. 
Figure~\ref{fig:pref-v-lat} compares the most related methods. Ours strikes the best latency and FPS improvement with the least image quality trade-off.

%% file: 02_related.tex
\section{Related Work}

In this section we discuss recent high-quality image generation methods and contrast with those that optimize runtime with a focus on face generation.

\parag{Photo-realistic synthesis.}
High-quality photorealistic rendering is booming, using either implicit scene representations such as Neural Radiance Fields (NeRFs) \cite{mildenhall2020nerf} or deep neural networks trained on GAN objectives~\cite{karras2018style,karras2020analyzing}, which can also be conditioned on viewpoint and pose changes~\cite{deng2020disentangled,tewari2020stylerig}. However, these implicit models all rely on very deep neural networks that do not run at high-enough frame rates or are limited to 
static scenes via pre-computed acceleration structures~\cite{yu2021plenoctrees}.

\parag{Talking head models.} For dynamic faces, it is most common to start from a parametric face model that parameterizes identity and expression in the form of blend shapes that are linearly combined with the full mesh~\cite{blanz1999morphable,li2017learning}. The coefficients of these models are low-dimensional and therefore suitable for driving avatars in telepresence and for reenactment by mixing identity and pose information of two subjects. While earlier models focus on only modeling the facial region~\cite{vlasic2005face,thies2015real,thies2016face2face,suwajanakorn2017synthesizing,thies2019deferred} conditioning on a complete head model, such as FLAME~\cite{li2017learning}, enables novel view synthesis of side views~\cite{ghosh2020gif} suitable for view-dependent 3D systems\cite{zhou20173dps}.  We follow this line of work and extend it by reducing the rendering latency for interactive applications.

Another line of work developed subject-agnostic face synthesis~\cite{olszewski2017realistic,nagano2018pagan,geng2018warp,fried2019text} that is conditioned on single or multiple images instead of a subject-specific 3D model. 
However, this comes at the price of reduced details and entanglement of expressions and head motion~\cite{wang2021one}, particularly if the reference and target poses differ largely, and may require explicit warping operations that are relatively costly and ill-defined for occluded regions~\cite{zakharov2020fast}. Although \cite{zakharov2020fast} is already extremely fast on medium image resolution with 4ms runtime, there is still room for improvement when rendering high-resolution images and medium resolution on embedded devices. 

\parag{Caching approaches.} Re-using information from the previous time step has been used for many computer vision tasks, among them: object detection~\cite{malinowski2020sideways}, video action recognition \cite{feichtenhofer2020x3d}, and segmentation \cite{hu2020temporally}.
Carreira et al.~\cite{carreira2018massively} gives an excellent overview of different architectures for video processing, including: 
\emph{i)} \emph{depth parallel} architectures that execute a deep neural network over several iterations, leading to a delay equal to the depth of the layers;
\emph{ii)} \emph{depth parallel + skip} in which the head (last couple of network layers) of a depth parallel network is updated with the new input via a skip connection; and,
\emph{iii)} \emph{multi-rate clock} architectures in which the input features for the head are not updated at every time step and the head and backbone operate at different clock rates. 
Our caching approach follows the multi-rate clock pattern.
However, none of the existing parallel models has demonstrated image generation.
Our contribution is to tailor this general concept to the problem of novel-view-synthesis of faces via neural rendering by a suitable form of warping.

%% file: 03_method.tex
\section{Preliminaries}

Our goal is an efficient neural renderer that outputs an image $\mI \in \R^{H \times W}$ of a face parameterized as a surface mesh with vertices $\vv \in \R^{3 \times K}$, triangle indices $\vi \in \R^{3 \times K}$, and associated texture coordinates $\vu \in \R^{2 \times K}$ and neural texture $\mN \in \R^{D \times H \times W}$ {\color{black}{\cite{thies2019deferred}}}. These are the same inputs that a normal renderer would expect, except that the texture $\mN$ is $D$-dimensional instead of storing three color values. 

\parag{Deferred neural rendering.} Our starting point is the deferred neural renderer introduced by Thies et al.~\cite{thies2019deferred}, which approximates the complex and computationally expensive rendering equation with a convolutional neural network $G$. Figure~\ref{fig:neural rendering} gives an overview including the differences of \cite{thies2019deferred} applied to our full model including caching.
Initially, a rasterizer renders UV maps $\mU \in \R^{2 \times H \times W}$ of the textured mesh. These are of the same dimension as the output image and store for every pixel the corresponding texture coordinate. Sampling these locations from $\mN$ gives the feature map $\mF \in \R^{D \times H \times W}$. For classical deferred rendering, we would sample from a color texture and combine it with light position information to form the final image. 
In the case of the neural renderer, the texture has more than three channels forming learnable features. 
The network $G$ turns $\mF$ into the final image, replacing geometric illumination computations and material shaders in classical rendering. This forward pass is traced with blue arrows in Figure~\ref{fig:neural rendering} while the backwards information flow during training is marked in green.

\parag{Training and face reconstruction.}
The parameters of the involved neural network $G$ as well as the neural texture itself are trained on a large dataset that has examples of the input 3D mesh and a high resolution image of the face---the desired output. 
We use real videos of the person as input and reconstruct vertices $\vv \in \R^{3 \times K}$, expression coefficients $\ve \in \R^{50}$, PCA shape parameters $\vs \in \R^{100}$, and head pose $\theta \in \R^{6}$ of the FLAME parametric model~\cite{li2017learning} alongside camera position $\vp \in \R^{3}$ using the off-the-shelf estimator DECA~\cite{DECA:Siggraph2021}. Internally, it is using the 2D keypoint detector from \cite{bulat2017far}. 
The reconstructed face overlays well with the image when re-projected, but details, such as hair and ears are often misaligned, which puts a larger burden on the neural renderer to synthesize these.
This reconstruction step is marked with a red arrow in Figure~\ref{fig:neural rendering}.

The generator $G$ is a U-Net and the %
neural texture $\mN$ is initialized at random and subsequently optimized by backpropagation to store details of the training object locally. The loss function is the L1 difference between the rendered and the reference image in the dataset. This backwards pass is traced with green arrows in Figure~\ref{fig:neural rendering}. Training is on cropped images, which speeds up training.

\parag{Base architecture.} We use the 10-layer U-Net as in \cite{pix2pix2017} and a multi-scale neural texture with four levels of detail as in \cite{thies2019deferred}. Furthermore, to better model viewpoint-dependent effects, the view direction is projected to 9 spherical harmonics coefficients which are subsequently multiplied to channels 4 through 12 of the feature map. This enables explicit encoding of view-dependent effects similar to positional encoding~\cite{vaswani2017attention}. 

\parag{Real-time viewpoint-dependent rendering.} Our primary application fields is 3D teleconferencing, where a person must be rendered at a high frame rate, with low latency, from the viewpoint of the user that is roughly frontal.
Given a new view direction, e.g., from an eye-tracker, our focus is on generating a natural looking image of this novel view as quickly as possible to mitigate motion sickness, reduce warping artifacts~\cite{zhou2017error}, and avoid discomfort. The object motion capture could be performed offline or through a slower channel, as usually only the viewpoint-dependent rendering demands the low-latency. In telepresence applications, 
bandwidth is dictated by the size of the FLAME model; estimated on the source side, transferred, and rendered by our system given a new view from the receiver.

\begin{figure}[t!]
\begin{center}
\includegraphics[width=0.8\linewidth]{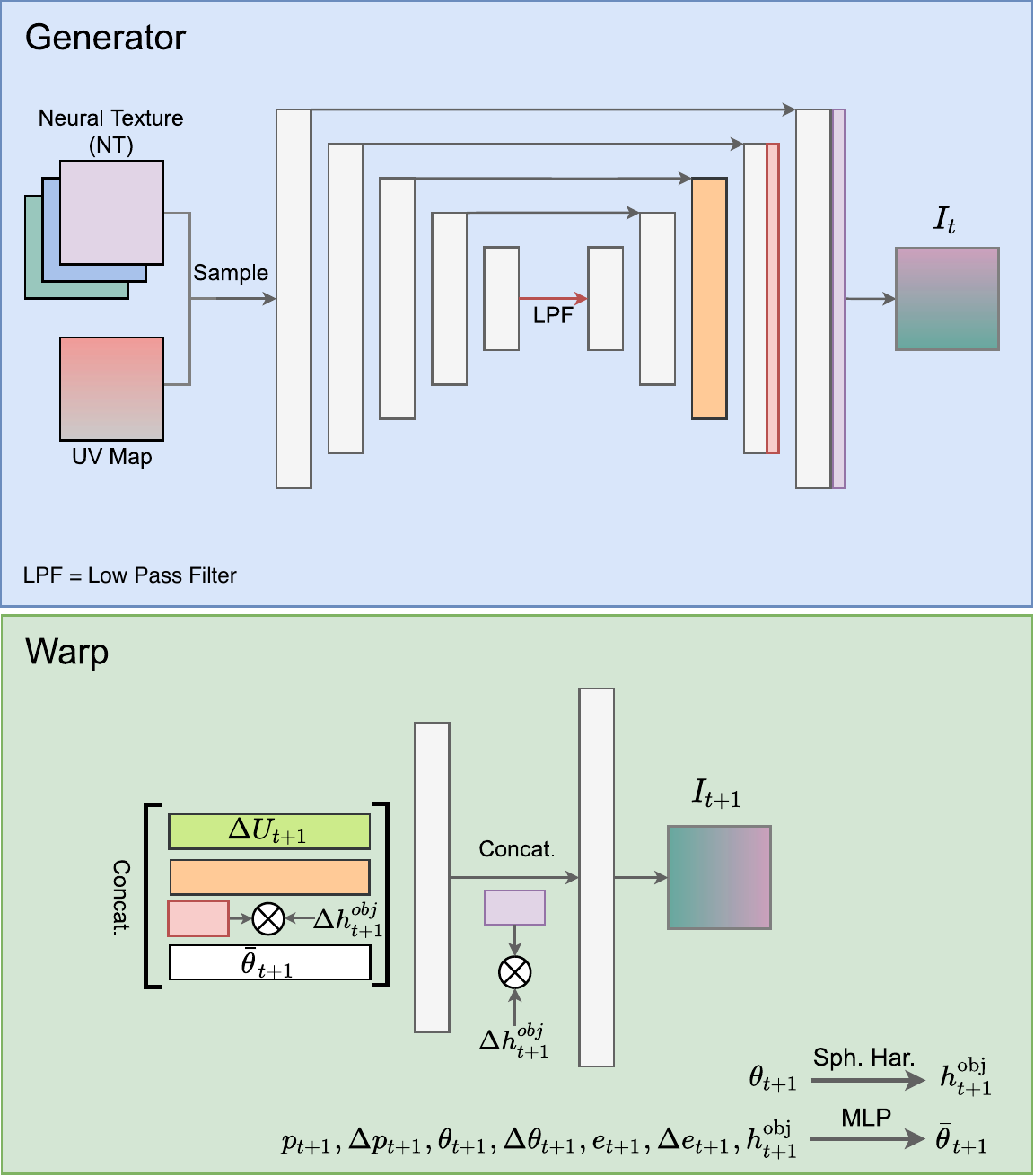}
\end{center}
\caption{\label{fig:warp_components}
		{\bf Our low latency pipeline} caches the previous frame (features from the generator (left) marked orange and red) and the pose of the person to generate the subsequent frames with only two additional up-convolution layers (right) as soon as a new view direction is available. The generator and warping networks are learned end-to-end and can be applied in parallel at inference time.
	}
\end{figure}

\section{Method} 

In this section, we introduce our neural caching approach, and propose two variants that operate on single and multi-GPU systems and are respectively tuned for latency and frame rate. 
We cache information from the immediately preceding frames. Thereby, the motion that must be bridged from the cached information to the current frame is small, which allows us to introduce an implicit warp that attains maximum performance.

\paragraph{Neural Cache.}
\label{sec:warping}
For our neural caching, we first run the deep and slow image generator $G(\mN_t,\mU_t,\allowbreak \vp_t)$ that is conditioned on the neural texture $\mN$, UV-map $\mU_t$ and view direction (e.g., user's head motion in VR) of the current frame~$t$. Figure~\ref{fig:warp_components} provides a detailed overview of our pipeline. 
We cache features $\mC^{(3)}_t$, $\mC^{(4)}_t$, and $\mC^{(5)}_t$ from the last 3 layers of the generator together with camera position $\vp_t$, and spherical harmonics (SH) encoding of the 
 pose, $\vh_t^\text{obj}$. Furthermore, we add the {\color{black}{UV map $\mU_t$}}, expression $\ve_t$, and 
the pose $\theta_t$ that are specific to rendering faces.
Formally we write the combined cache $\mC$ as

\begin{equation}
    \mC_t := [\mC^{(3)}_t, \mC^{(4)}_t, \mC^{(5)}_t, \theta_t, \vp_t, \ve_t, \vh_t^\text{obj}, {\color{black}{\mU_t}}].
\end{equation}

\parag{Implicit Warping.} 
Previous work used an explicit warp operation from a reference frame, which requires preceding neural network layers to predict the warp and is implemented as a texture sampling step that is relatively slow due to random access patterns for every pixel. 
We propose an implicit warp with a neural network $W$ that is composed of only two up-convolution layers. These two layers take in cached information $\mC_t$,
rendered UV map and 
the new camera, pose, and expression from the new frame $t+1$,
to reconstruct the image $I_{t+1}$. This network is intentionally kept shallow to decrease the latency of image generation.

In addition, we found in a detailed ablation study that giving as input the new camera position $\vp_{t+1}$, object pose $\theta_{t+1}$, expression $\ve_{t+1}$ and their differences to the previous frame, via a single-layer MLP, $M$, works best. This yields
\begin{equation}
    \bar{\theta}_{t+1} = M(\vp_{t+1}, \Delta \vp_{t+1}, \theta_{t+1}, \Delta \theta_{t+1}, \ve_{t+1} \Delta, \ve_{t+1}, \vh_{t+1}^\text{obj}),
\end{equation}
where the $\Delta$ refers to the change in a quantity between two frames, here $\theta_{t+1} - \theta_{t}$. {\color{black}{Additionally, we also include the UV map $\mU_{t+1}$.}}
Together with the cached features processing by the 
first warping layer $W_1$ gives
\begin{equation}
    \mF_1 = W_1(\mC^{(3)}_t, \mC^{(4)}_t \Delta \vh_{t+1}^\text{obj}, \bar{\theta}_{t+1}, \Delta \mU_{t+1})
\end{equation}
where $\Delta \vh_{t+1}$ and $\bar{\theta}_{t+1}$ are broadcasted to the resolution of the cache and $\Delta \vh_{t+1}$ modulates the feature map by multiplication, similar to how positional encoding works, but here for rotational changes.
These features are further processed to output the image with
$I_{t+1} = W_2(\mF_1, \mC^{(5)}_t \Delta \vh_{t+1}^\text{obj})$.
Note that $W$ warps the previous to the current frame, but without a spatial transformation layer---it does so implicitly via a local approximation of how the image changes with respect to changes in pose.

\begin{figure}[t!]
\begin{center}
\includegraphics[width=0.8\linewidth]{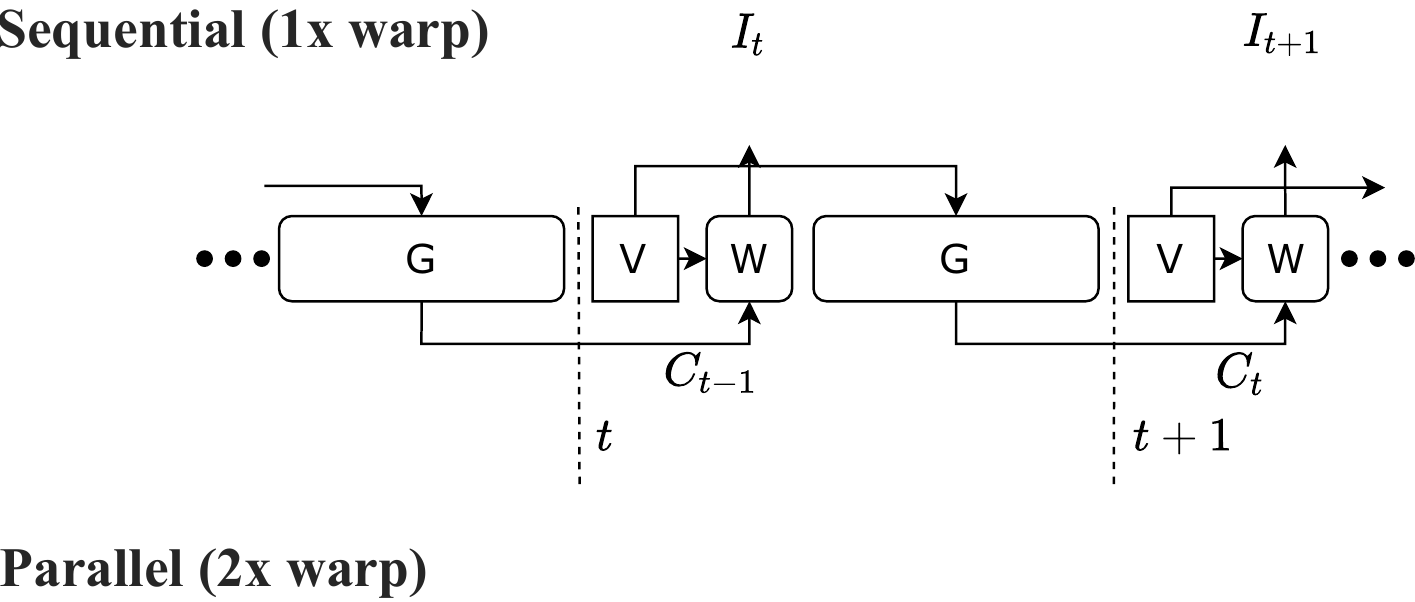}
\includegraphics[width=0.8\linewidth, trim={0cm 19.5cm 2cm 0cm},clip]{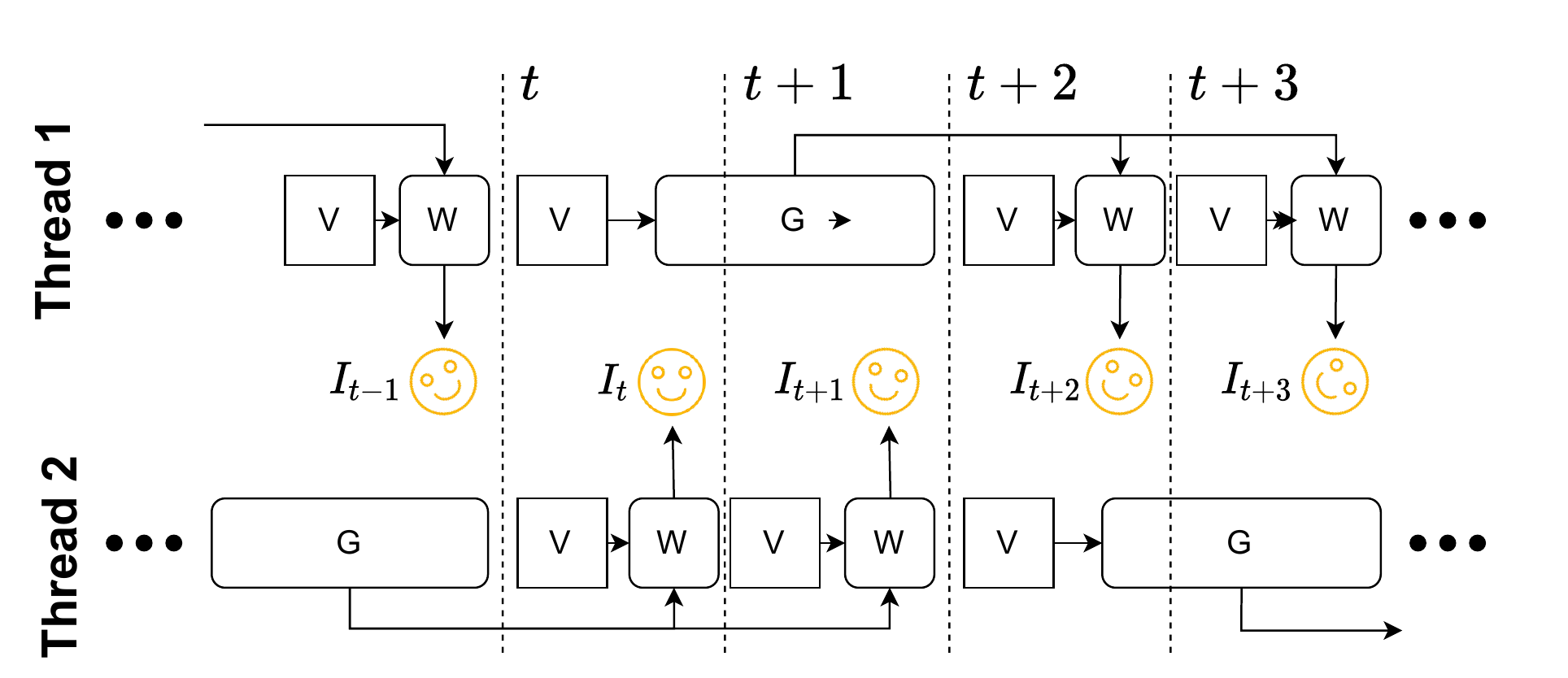}
\end{center}
\caption{\label{fig:latency}
		{\bf Our rendering pipelines for sequential (Top) and parallel execution (Bottom).} A network $W$ warps a cache from the previous frame as soon as a new viewpoint $V$ is ready for rendering, producing the output image much quicker than the full generator $G$ could. Leveraging two threads further increases the output frame rate.
	}
\end{figure}

\subsection{Operation Modes and Parallel Execution}

Our approach scales easily between single and parallel GPU execution. When only a single GPU is available, Figure~\ref{fig:latency}-top visualizes the principle of reducing latency with a shallow warp network, compared to running the slower generator. Here, the cache is  updated in sequence with the warp, thereby reducing latency but not frame rate.

When multiple GPUs are available, we can combine the proposed implicit warp with parallel execution, thereby rendering at a two or more times higher frame rate %
by warping a single or multiple images $\mI_t,\mI_{t+1},\cdots$ while the generator runs on a separate GPU on a separate thread. Figure~\ref{fig:latency}-right showcases this scheme. Notably, even though the warping is an approximation, multiple warping does not reduce image quality when operating online on high-frame rate video streams. The higher processing speed of warping twice in parallel reduces the distance between frames that can be processed and thereby makes the two-frame warp as difficult as a single-frame warp operating at half the frame rate.

We experimented with different job assignment and synchronization schemes between the two threads. We found that threads alternating between image generation and warping is most efficient and eases implementation. In this model, the main thread distributes newly arriving viewpoint information to the two worker threads via a queue, each associated to one GPU. These in turn wait for new data in the queue. Upon receiving new data, they alternate between caching and warping as visualized in Figure~\ref{fig:latency}-bottom. Thereby, the warping is executed on the same GPU as the generator, such that the cache can remain on the same GPU. The alternative of using a dedicated warp and cache thread has a much lower performance since the cache would have to be moved from one GPU to CPU and then again from CPU to the target GPU.
The required synchronization of ques has a negligible overhead in our implementation on two RTX 2080s with only 0.25ms/frame.
Moreover, our preliminary attempt of using manual locks instead of queues was more complex without improving performance. We will make our implementation publicly available to facilitate further research.

\subsection{Improved Neural Head Rendering}
\label{sec:architecture}

Our starting point, DNR~\cite{thies2019deferred}, is a general rendering approach. In the following, we explain our architectural changes towards tuning it for face synthesis.

\parag{Head stabilization}
To reduce jitter and flicker of the head, 
we ensure that the virtual camera that we use to generate our UV masks is always centered on the subject's head and has consistent scale when generating videos by centering the camera on the head midpoint and scaling by the projected ear-to-ear distance.
Because the driving motion capture signal is often unstable, we further smooth the global head position $\vp_t$ with a delayed Gaussian filter of size five and fix the identity $\vs$ of the FLAME model to be the mean identity estimated by DECA on the training set. To maintain facial expressions and lip motion faithfully, the jaw orientation in $\theta$ and expression parameters $\ve$ are left untouched.

\parag{Loss Function} For our baselines which only predict the image at time $t$ we define the loss function as 
\begin{equation}
    L_\text{train} = \lambda_\text{tex} L_\text{tex} + \lambda_\text{img} L_\text{img} + \lambda_\text{p} L_\text{p} ,
\end{equation}
which measures the L1 distance between the first three channels of the sampled neural texture and the ground truth image ($L_\text{tex}$), the L1 photometric loss ($L_\text{img}$), and the perceptual loss \cite{johnson2016perceptual} between the predicted image and ground truth image ($L_{p}$). We weight these loss terms with coefficients 1, 1, and 0.1. For the warping we add $\lambda_\text{img}$ and $L_\text{p}$ on the future frame ${t+1}$ and down-weight the existing $L_\text{img}$ by 0.1 to put much more weight on the prediction for $I_{t+1}$ that is used at inference time.

\parag{Architectural Changes.} Based on the work of \cite{karras2021alias} we make the following improvements to the base DNR \cite{thies2019deferred} network to improve output image quality. First, we replace the transpose convolutions in the latter half of of the U-Net with a bilinear upsample layer followed by a 2D convolution (up-convolution). This has been shown to increase the final image quality of output and reduce grid-like noise in reconstructed images. {\color{black}{Furthermore, we apply a Gaussian low pass filter (LPF) on the smallest spacial features of the U-Net architecture.}}
We refer to the baselines that utilize all these improvements as DNR+.

%% file: 04_experiments.tex
\section{Experiments}
\label{sec:Experiments}
We evaluate our Neural Warping technique with respect to our goal of maximizing image quality and minimizing the latency by applying our Neural Warping. Figure~\ref{fig:pref-v-lat} summarizes our main results on the possible trade-offs between accuracy vs.~latency and accuracy vs. fps, comparing our two variants to the most related work and showing an improvement in fps of up to $300\%$ and a reduction in latency of $70 \%$. We also provide an ablation study to identify specific trade-offs with respect to individual components of our warping network. The supplemental material contains additional results.

\parag{Baselines.}
\textbf{DNR}~\cite{thies2019deferred} is the backbone we use as our reference.
\textbf{DNR+} improves DNR with recent neural network architecture improvements from~\cite{karras2021alias}. We furthermore add \textbf{DNR++} that uses the current and past frame as input. DNR+ and DNR++ both act as a theoretical upper bound for our model's image quality and therefore, provide a good measure of the effectiveness and efficiency of our approach.
We also compare against the recent method from \textbf{Wang et al.}~\cite{wang2021one}, using their online interface (\href{http://imaginaire.cc/vid2vid-cameo/}{http://imaginaire.cc/vid2vid-cameo/}) and in terms of runtime to \textbf{Zakharov et al.}~\cite{zakharov2020fast} using the same $512\times512$ image resolution.
We do not provide PSNR numbers as it was designed for a much smaller resolution of $256\times256$.
In addition, we create a \emph{naive} baseline where we shift input frames by one to emulate the delay that incures when running a large image generator without warping. It serves as an expected lower bound on accuracy.

\parag{Datasets.}
We use the talking head sequences from \cite{thies2019deferred} to compare against prior work as well as a \emph{beard} and a \emph{high-fps} sequence with more difficult facial hair that was recorded with ethics board approval. 
Videos for the \textbf{\emph{beard}} and \textbf{\emph{high-fps}} dataset were recorded at 1920x1080 and split into, respectively, 2604/558/558 and 3600/500/1000 frames for training/validation/testing. \textbf{\emph{Trump}}~\cite{thies2019deferred} sequences is a 1280x720 of 431 frames.
The \textbf{\emph{Obama}}~\cite{thies2019deferred} sequence is a 512x512 of 2412 frames. 
The \textbf{\emph{male}} and \textbf{\emph{female}}~\cite{thies2019deferred} sequences are both 768x768 with 2380 frames. The \emph{high-fps} recording has 60 fps, all others run at 30 fps.

\parag{Training Setup.} We train all our models for 150 epochs using Adam~\cite{kingma2014adam} with betas equal to \{0.9, 0.999\}, learning rate of 1e-4 for both the Generator, $G$ and Warp network $W$, and 1e-3 for the neural texture.

\begin{figure}[t!]
\begin{center}
\includegraphics[width=0.85\linewidth]{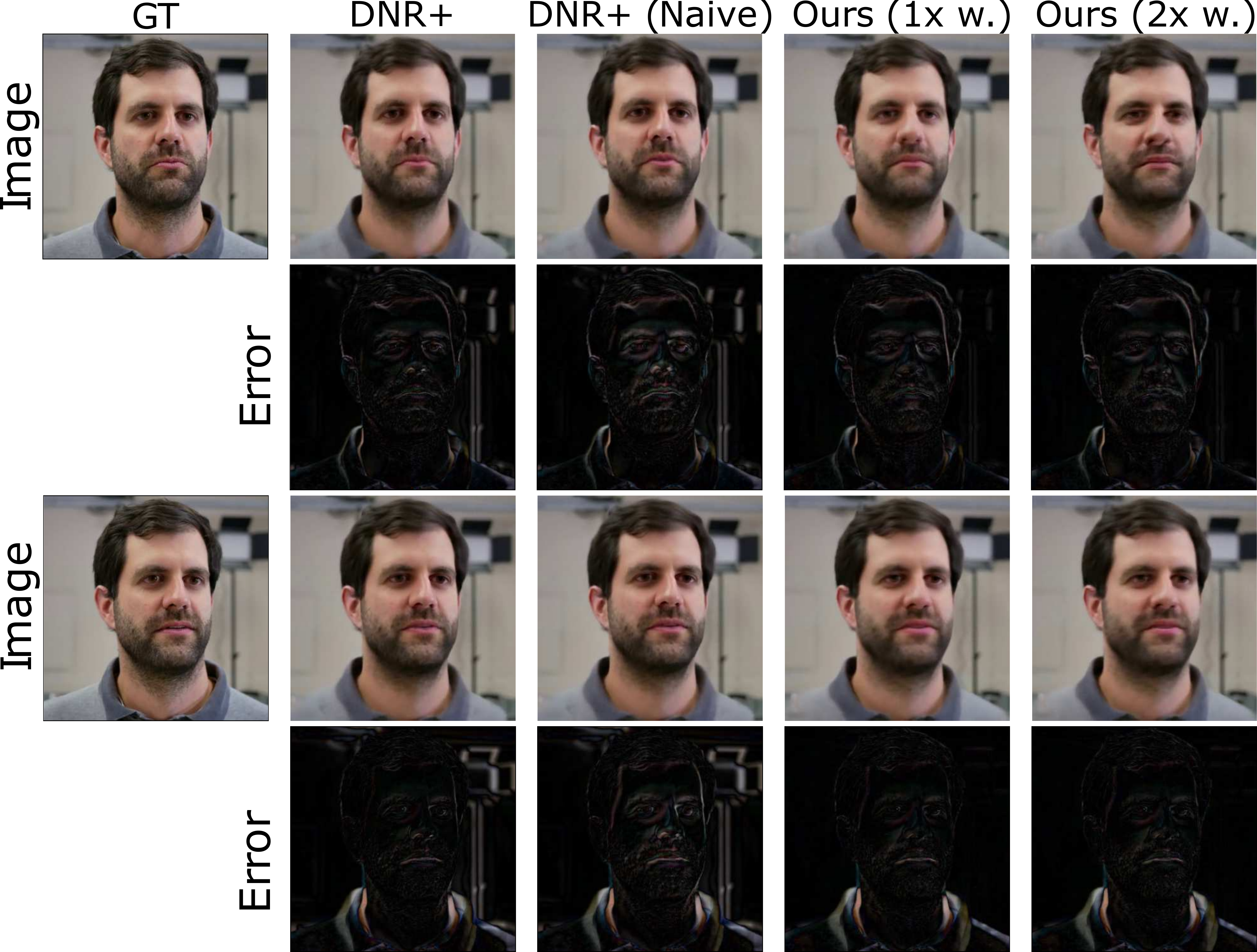}
\end{center}
\caption{\label{fig:errormap}
		{\bf Frames and their error maps} for our model (1x and 2x warp), DNR+, and naive baseline. Based on the L1 error we can see all models perform similarly, differences are visible in the error maps for high-frequency details.}
\end{figure}

\parag{Metrics.} %
We evaluate our image reconstruction accuracy using an L1 reconstruction error between the ground truth and reconstructed image as well as PSNR and SSIM. Furthermore, to show our increase in speed, we report the latency and frames-per-second (fps) of our models and baselines. For all methods, we only measure the time it takes to process the input UV maps and the corresponding skeleton (pose), expression, and camera (extrinsics) information; excluding the time it takes to render the UV map  because they are implementation dependent and with negligible overhead when implemented in a rasterizer. Similarly, we only account for the time the inference generator takes in \cite{zakharov2020fast}, ignoring the processing of the conditional keypoint image. All latency and fps metrics are computed on NVIDIA RTX 2080 GPUs. For our approaches we report metrics and timings when we cache every frame (1x warp) and every second frame (2x warp) to show our method is robust to varying deviations between input frames.

%% file: 05_results.tex
\subsection{Latency and Runtime Improvement}

The generator backbone has a runtime of 47.02ms and equivalent latency. Our warp net has a runtime of 14.62ms and therefore deduces latency by a factor of 3.2. Running on multiple GPUs improves frame rate from 28.5 to 67.6 while inducing only a negligible 0.25 ms increase in latency due to the required synchronization. Note that parallel execution across multiple GPUs can by itself not improve the latency of a sequential process. When processing even and odd frames on different GPUs, the time from input view to rendering output remains the same. The impact of different model configurations on latency is evaluated in Table~\ref{tab:ablation_quant}.

\subsection{Offline Reconstruction Quality}

\input{tables/main_quant}

To test image generation quality, we use a held-out test video and drive the trained models using the FLAME head model reconstructed on the reference video.
The results from Figure~\ref{fig:errormap} and Table~\ref{tab:quant_main} evaluate the difference to the reference video in offline processing mode.
When comparing individual frames between the baseline and our model, Figure~\ref{fig:errormap} shows that single (\VOOne{}) and multi-frame warping (\VTTwo{}) work nearly as well in terms of error. Table~\ref{tab:quant_main} reveals, somewhat surprisingly that \VOOne~ slightly outperforms the DNR+ baseline in terms of PSNR despite having to warp with a shallow network. This is possible since it has access to the current and past frame that helps to correct errors in the facial expression estimation. To this end, we introduced the DNR++ baseline as a new upper bound that also has access to the two previous frames. In summary, at a small reduction in accuracy compared to the baseline models, latency and frame rate are greatly improved by a factor of two or more.

\subsection{Online Reconstruction Quality}

\begin{figure}[t!]
 \begin{center}
 \includegraphics[width=0.95\linewidth, trim={0cm 0cm 0cm 0cm},clip]{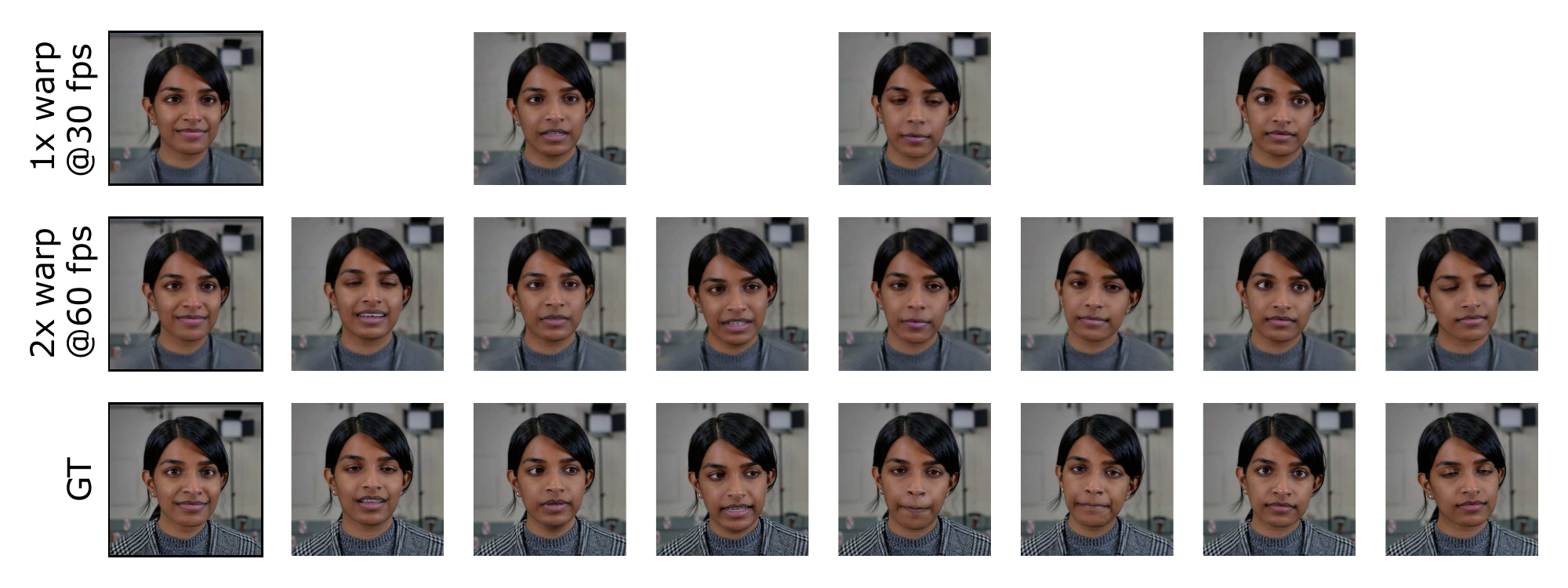}
 \end{center}
 \caption{\label{fig:qualitative_highfps}
 		{\bf Online application.} When running at their native frame rate, high-fps models (\VTTwo~) improve as they have to bridge a smaller gap between frames.
 	}
 \end{figure}

Online reconstruction requires the algorithm to run at the native frame rate of the video. This has a significant influence on the performance of our algorithm as the warping operation becomes simpler for high-frame rate videos where the motion between two frames is reduced, leading to even higher performance gains than for the previous offline evaluation. We test this effect on the \emph{high-fps} sequence shown in Fig.~\ref{fig:qualitative_highfps}. Table~\ref{tab:ablation_highfps-compare} reveals that running online on 60 fps videos with Ours (2x warp) improves on Ours (1x warp), as the latter can only process every other frame requiring larger warps. Hence, warping multiple times is beneficial, jointly improving in latency, runtime, and image quality, when parallel hardware is available. The basic DNR baselines do not even run at 30 fps and are hence not comparable in the high-fps online setting.

\input{tables/highfps_compare}

Note that absolute PSNR numbers differ across subjects and scenes since faces are smaller/bigger and also contain more or less high-frequency details, including the texture on the shoulder region that is of little concern for the facial feature reconstruction fidelity. Hence, it does not make sense to relate absolute but only relative numbers between the \emph{beard} and \emph{high-fps} scores.

\subsection{Novel View Synthesis Quality}
\label{sec:novel-view}
\begin{figure}[t!]
\centering
  \centering
    \includegraphics[width=0.8\linewidth, trim={0cm 0cm 0cm 0cm},clip]{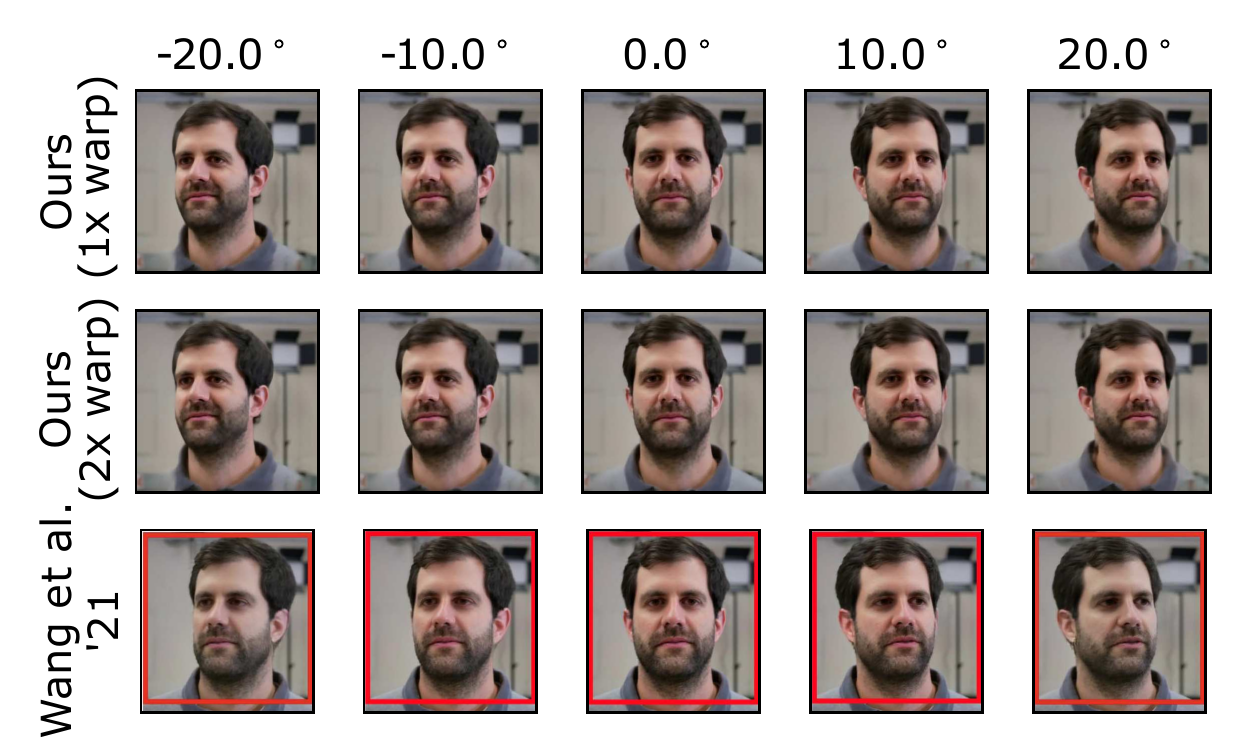}
  \caption{\label{fig:nvidia}
		{\bf Comparison against Wang et al. \cite{wang2021one}}. While both models show similar levels of detail, ours is anchored in a 3D representation which gives us more fine-grained and independent control.
	}
\end{figure}

\begin{figure}[t!]
  \centering
    \includegraphics[width=0.8\linewidth, trim={0cm 0cm 0cm 0cm},clip]{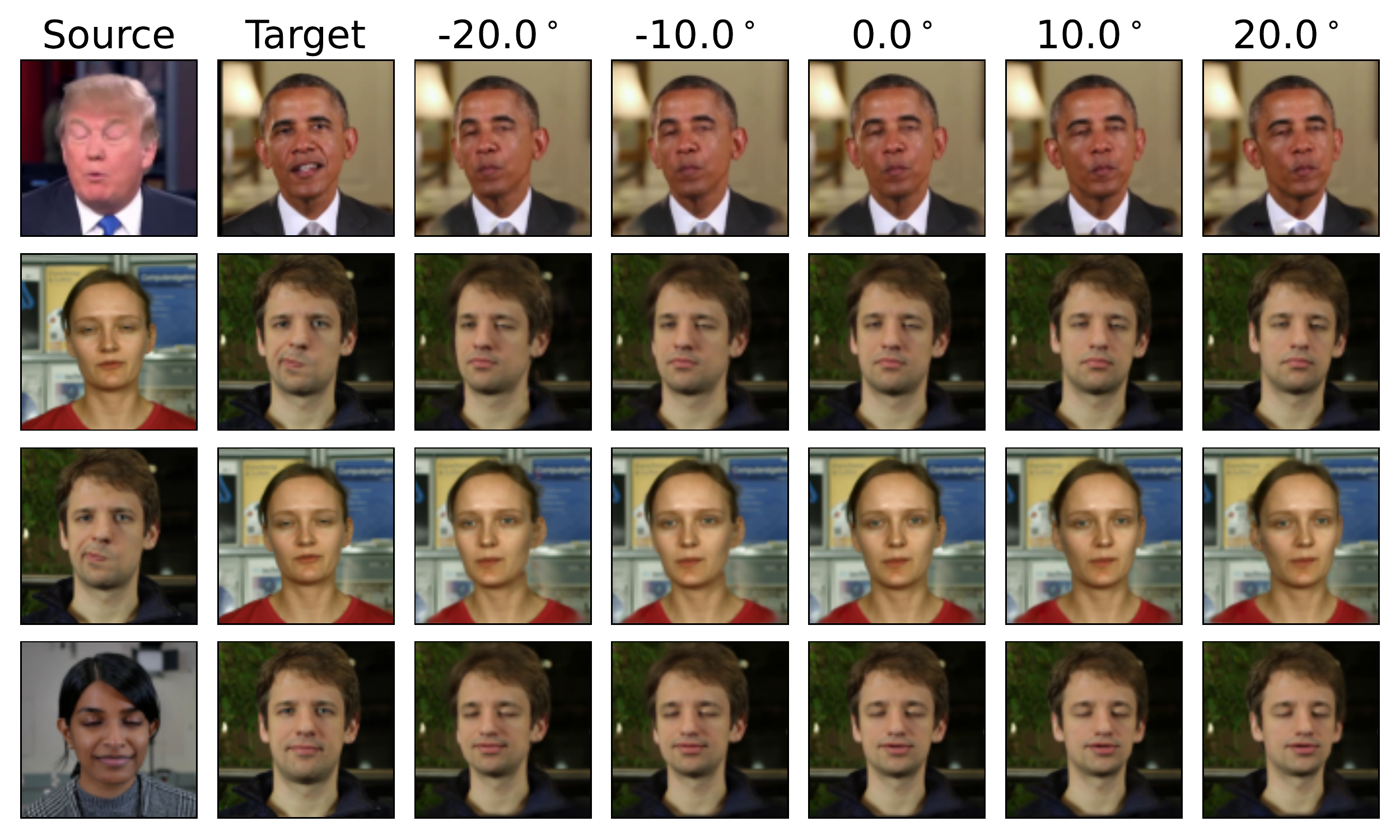} 
    \caption{\label{fig:retarget_novel}
		{\bf Novel views while retargeting}. Our model is capable of generating realistic, view-dependent novels views {\color{black}{(Section~\ref{sec:novel-view})}} while mimicking the source's facial expressions {\color{black}{(Section~\ref{sec:retarget})}}. These examples are generated using Ours (\VOOne~) using the target actors from \cite{thies2019deferred}.
	}
\end{figure}

Reproducing a pre-recorded sequence does not necessarily need low latency since the entire video could be cached. Yet, latency is crucial for rendering a face from a novel viewpoint to account for the user's head motion in VR and in general for viewpoint-dependent displays. We generate novel views of characters by rotating their underlying 3D mesh (which is used to generate our input UV maps) while holding other parameters fixed. Figure~\ref{fig:retarget_novel} shows the retargeting of poses and views across videos and \ref{fig:nvidia} shows synthetically generated views and compares it to the results from \cite{wang2021one}. Because we condition on a full 3D face model, our rotation is more precise and keeps the pose unchanged compared to the learned 3D features from Wang et al.~that lead to opening of the mouth and up-rotation. The quality by \cite{wang2021one} is expected to be slightly lower as it is not person-specific, which serves our motivation to train a person-specific model.

\subsection{Retargeting} 
\label{sec:retarget}
\begin{figure}[t!]
\begin{center}%
\includegraphics[width=0.85\linewidth,trim={0cm 0.1cm 0cm 0cm},clip]{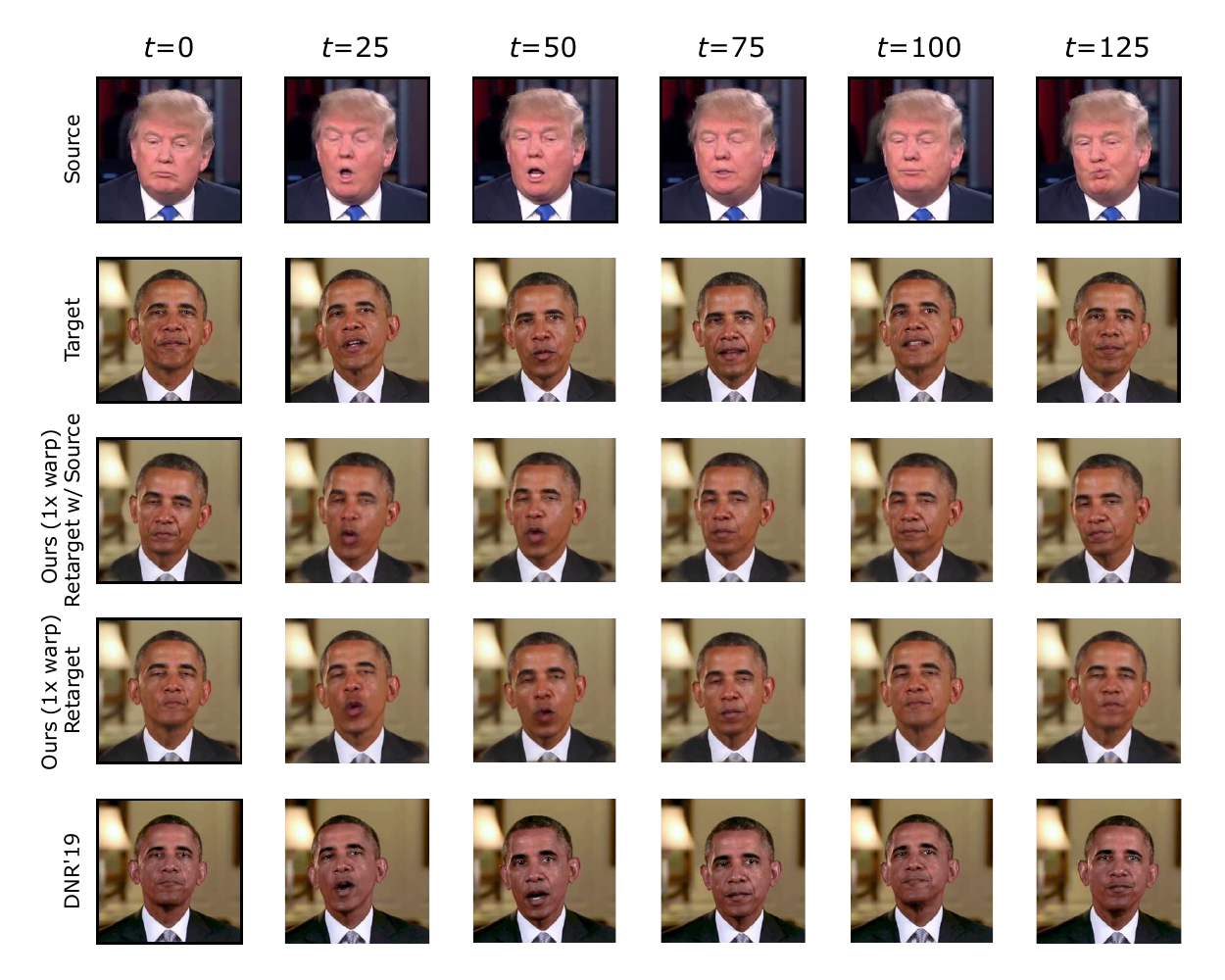}%
\end{center}%
\caption{\label{fig:retarget_main}
		{\bf Retargeting}, without and with transfer of the global head motion, including the comparison to~\cite{thies2019deferred}.}
\end{figure}

To show the flexibility of our approach, we retarget facial and head motions from one person to another on in-the-wild videos that were established as a benchmark in \cite{thies2019deferred}. In this setting we train the neural texture and renderer on our target subject and use these learned models at inference time, driven by the head motion reconstructed from a source sequence with a different actor. In Figure \ref{fig:retarget_main} we show that our approach is capable of retargeting with just transferring expressions (as in \cite{thies2019deferred}) and also when mapping the global head orientation. The original results, provided from \cite{thies2019deferred} do not include head stabilization. Nevertheless, this comparison shows that our approach (\VOOne~) is reproducing or even outperforming their image quality.

\input{tables/ablation_quant}

Furthermore, as we condition on a full 3D head model, we are able to generate novel views while performing the retargeting as shown in Figure~\ref{fig:retarget_novel}. Since our approach only approximates the background for each scene, we use a static background in our predictions.

\subsection{Ablation Study}

\parag{Network Backbones.}
To show our method's generality to other backbones, we compare our UNet to the widely successful ResNet-backbone using residual blocks. As expected, warping is effective in reducing latency from 113 ms to 26 ms, but the ResNet does not outperform the UNet architectures. On the \emph{beard} sequence, the ResNet backbone operating using single warps gives a PSNR of 23.70 (UNet 26.66) and 23.18 PSNR using two warps (UNet 26.45).

\parag{Model Components.} Adding the parameters and transformations used in \emph{Our} full model one-by-one increases image quality (L1, PSNR, and SSIM) while only incurring a slight decrease in latency and fps (see relative latency and relative fps column, the relative latency compared to our full model). Table~\ref{tab:ablation_quant} presents the results for the sequential operation mode on the \emph{beard} sequence over the following components:
\begin{itemize}[noitemsep]%
\item{\bf Concat~UV:} using the difference of the current and cached UV maps in $\mC_3$.

\item{\bf Use $\mathbf{\theta}$:} concatenating the cached pose and camera extrinsics to $\mC_3$.

\item{\bf Use MLP:} passing $\theta$, $\ve$, $\vp$ through an MLP $M$ before concatenation with~$\mC_3$.

\item{\bf SH Pose:} includes the spherical harmonics $S^{obj}$ in $\theta$.

\item{\bf SH Skips:} 'rotationally' encode $\mC_4,\mC_5$ using $\Delta \mS^{obj}$.

\item{\bf ExWarp:} explicit warping from a reference by sampling the learned neural texture with $\mU_{t+1}$ and concatenating it with $\mC_3$.

\item{\bf Exp:} cache and concatenate the expression $\ve$ with $\theta$.%
\end{itemize}%
The explicit warping (\emph{ExWarp}, second-last row) adds a large latency increase while not improving the quality metrics consistently. Hence, we favour the implicit warp in our full model. When applied in parallel on multiple GPUs, these reduced latencies translate directly to improved frame rate. The synchronization overhead in the parallel implementation is only 0.25ms/frame, which we measured by running the sequential model with the same threading and queue synchronization as for the parallel mode and taking their latency difference.

%% file: tables/main_quant.tex
\begin{table}[b!]
\centering
\resizebox{1\linewidth}{!}{%
\begin{tabular}{l|c|ccc|cc}
Model       & \#GPU         & L1 $\downarrow$              & PSNR $\uparrow$           & SSIM $\uparrow$           & Latency {[}ms{]} $\downarrow$ & FPS $\uparrow$          \\ \hline \hline
timing baseline \cite{zakharov2020fast} & 1 & -               & -              & -               & {16.60}       & {60.2}    \\ \hline
DNR+ (Naive)         & 1 & 0.0278          & 25.56          & 0.8970           & 46.81            & 21.4          \\
DNR+                 & 1 & \underline{0.0240}     & {26.59}    & \underline{0.9165}    & 46.81            & 21.4          \\
DNR++                & 1 & \textbf{0.0237} & \textbf{26.67} & \textbf{0.9168} & 49.37            & 20.3          \\
\hline
ExWarp as in \cite{zakharov2020fast} (1x warp) & 2   & 0.0257          & 26.33          & 0.9108          & 18.84            &  26.4        \\
ExWarp as in \cite{zakharov2020fast} (2x warp) & 2   & 0.0260          & 26.24          & 0.9094          & 18.84            & \underline{60.8}         \\
\hline
Ours (1x warp) & 1  & 0.0244          & \underline{26.66}          & 0.9107 & \textbf{14.62} & 16.3 \\
Ours (1x warp)& 2             & 0.0244          & \underline{26.66}          & 0.9107          & \underline{14.87}    & 28.5         \\
Ours (2x warp)& 1  & 0.0251          & 26.45          & 0.9069 & \textbf{14.62} & 26.3 \\
Ours (2x warp) & 2            & 0.0251          & 26.45          & 0.9069          & \underline{14.87}   & \textbf{67.6}
\end{tabular}
}
\caption{\label{tab:quant_main}
		{\bf Offline evaluation on the $\emph{beard}$ dataset.} As expected, our model does not achieve the best metrics on the image reconstruction metrics, but they outperform the baselines in terms of latency and fps. Timing results for \emph{Ours 2x warp} are using parallel execution.}

\end{table}

%% file: tables/highfps_compare.tex
\begin{table}[b!]
\centering
\resizebox{.8\linewidth}{!}{%
\begin{tabular}{l|ccc}
Operation Mode            & L1 $\downarrow$             & PSNR $\uparrow$           & SSIM $\uparrow$           \\ \hline \hline
Ours (\VOOne~ @ 30fps)  & 0.0361          & 23.19          & 0.8148         \\
Ours (\VTTwo~ @ 60fps) & \textbf{0.0359} & \textbf{23.23} & \textbf{0.8162}
\end{tabular}
}
\caption{\label{tab:ablation_highfps-compare}
		{\bf \emph{high-fps} comparison} for our warping network using realistic operation settings on the \emph{high-fps} dataset. \VTTwo~ not only improves speed but even slightly improves the rendering quality.}

\end{table}

%% file: tables/ablation_quant.tex
\begin{table*}[t!]
\centering
\resizebox{0.7\linewidth}{!}{%
\begin{tabular}{ccccccc|ccc|cc}
Concat UV & Use $\Theta$ & Use MLP & SH Pose & SH Skips & ExWarp & Exp. & L1 $\downarrow$             & PSNR $\uparrow$            & SSIM $\uparrow$           & Latency {[}ms{]} $\downarrow$ & Rel. Latency {[}ms{]} $\downarrow$  \\ \hline \hline
 $\checkmark$      &      &         &         &          &        & & 0.0276          & 25.6799          & 0.8987          & 12.80            & -                       \\
 $\checkmark$      &  $\checkmark$ &         &         &          &        & & 0.0275          & 25.6994          & 0.8989          & 13.24            & 0.44                     \\
 $\checkmark$      &  $\checkmark$ &  $\checkmark$    &         &          &        & & 0.0271          & 25.8378          & 0.9012          & 13.61            & 0.81                      \\
 $\checkmark$      &  $\checkmark$ &  $\checkmark$    &  $\checkmark$    &          &        & & 0.0260          & 26.1627          & 0.9038          & 13.79           & 0.99                      \\
 $\checkmark$      &  $\checkmark$ &  $\checkmark$    &  $\checkmark$    &  $\checkmark$     &        &  &0.0244 & 26.5705 &  0.9105    &  14.40            & 1.60                             \\ 
  $\checkmark$      &  $\checkmark$ &  $\checkmark$    &  $\checkmark$    &  $\checkmark$     &  $\checkmark$   & & {0.0257}    & { 26.3326}    & \textbf{0.9108} & 18.84            & 6.03           \\        
 \hline
$\checkmark$      &  $\checkmark$ &  $\checkmark$    &  $\checkmark$    &  $\checkmark$     &  & $\checkmark$ & \textbf{0.0244} & \textbf{26.6562} & {0.9107} & 14.62 & 1.82 \\
 \end{tabular}
}
\caption{\label{tab:ablation_quant}
		{\bf Ablation study results} for our warping network using the sequential variant (1x warp), showing the relative improvement resulting from including each component on the \emph{beard} dataset.} 

\end{table*}

%% file: 06_discussion.tex
\section{Limitations}

Because our generator, $G$, and warping network, $W$, learn how to generate an image without an explicit rendering equation, we require a diverse set of training views to ensure that we can perform novel view synthesis and accurate warping at inference time. We can see in Figure~\ref{fig:retarget_novel} that the novel views break at extreme angles around the ears of the target subject as these are not seen during our training videos. The limiting factor is here the face reconstruction algorithm that becomes unreliable when large parts of the face are occluded. Eyes and a wide-open mouths can also pose a problem since they are represented as holes in the underlying FLAME model and therefore the direction of the eye gaze and tongue cannot be modeled. Furthermore, because the FLAME model only estimates 3D models for the head, the network struggles in cases where users have glasses or expressions that it cannot express. Improving image quality in these directions is largely orthogonal to our contributions towards low latency and high frame rate.

%% file: 07_conclusion.tex
\section{Conclusion}

We introduced an implicit warping method that reduces the latency and, if parallel hardware is available, increases the frame rate of neural face rendering. We believe that such parallel execution to reduce latency and increase frame rate will gain importance with VR and AR emerging on the consumer market at scale, as our caching approach is compatible with the deeper neural network architectures required to meet the ever-increasing demand for output resolution. Thus, our work makes an important step towards end-to-end VR and 3D telepresence using view-dependent displays.

\paragraph{Acknowledgements} This work was sponsored in part by the Natural Sciences and Engineering Research Council (NSERC), Huawei, and Compute Canada. We also want to thank Justus Thies for providing their dataset and baseline.

%% file: 08_supplemental_arxiv.tex
\appendix
\section*{Appendix}
\section{Implementation details}

Meshes are generated using the forward function of the FLAME model implemented in python~\cite{li2017learning}. UV maps are generated by rendering the FLAME head model using the PyTorch3D rasterizer, which implements rasterization on the GPU sing custom cuda kernels but is not as fast as classical rasterization.
The explicit warp is implemented using the pytorch \emph{grid\_sample} function, which is the same as used by \cite{chu2021fast} for their explicit warp.

\paragraph{Generating Novel Views}
Our method relies on both the processed UV maps, initialized with the 3D mesh estimated by DECA \cite{DECA:Siggraph2021}. Therefore, to generate novel views for our method we require to modify these parameters. In our tests, we found that the best generalization to novel views is obtained by rotating the vertices outputed by the FLAME model, as modifying the joint angles in FLAME model to extreme poses may lead to parameter ranges not seen during training.

\paragraph{Multithreading Library} 
Our multi-gpu operation modes are implemented using the PyTorch multithreading library and two threads. For synchronization and communication between threads we use the \emph{torch.multiprocessing.mp.Queue} ques: one input queue for each thread (two for two GPUs), one image output que, and one que to signal completion of caching. Listing~\ref{algo:parallelAlgorithm} provides pseudocode for how caching and warping threads are orchestrated from the main thread, making sure that the output images arrive in sequence and that the caching completes before warping on the same worker while also ensuring that the parallel workers are independent otherwise for maximal performance. For simplicity, the pseudocode is simplified for 2x GPU and 2x warping for every caching operation. A different number of GPUs and warp operations only requires to change the condition in line 4 and line 8 that distribute caching and warping operations over worker threads. 
We will publish our multithreading code so others can build upon it with their own neural models.

\begin{algorithm*}[]
    \DontPrintSemicolon 
    \SetKwBlock{DoParallel}{do in parallel}{end}
   
    \tcc{Main thread} 
    Queue viewpoint\_queue[NGPU=2], cache\_queue, out\_queue \tcp*{Queues used for synchronization}
    cache\_tid $\leftarrow$ 0 \tcp*{Pointer to assign worker responsible for caching}
        \For{t=0; t++;}
        {
            \KwIn{viewpoint, expr}
        \If{t \% NUM\_WARPS == 0}{
            cache\_queue.get() \tcp*{Wait for caching to complete}
            cache\_tid $\leftarrow$ (cache\_tid+1) \% NGPU \tcp*{Switch role of threads}
            \tcp{Post viewpoint to worker and assign caching role}
            (viewpoint, expr, warp=False) $\rightarrow$
            viewpoint\_queue[cache\_tid] 
         }
         \tcp{Post viewpoint to the other worker and assign warping role}
        (viewpoint, expr, warp=True) $\rightarrow$
        viewpoint\_queue[(cache\_tid+1) \% NGPU] \;
        \tcp{Display output image once available}
   \KwOut{out\_queue.get()}%
        }
    
    \tcc{Worker thread, one for each GPU}
    
    \DoParallel{
    NeuralCache cache \tcp*{variable local to each thread}
        (viewpoint, expression, warp) $\leftarrow$ viewpoint\_queue.get() \tcp*{Wait for new input to process }
        \If{warp = True}{
            image $\leftarrow$ runWarpNet(viewpoint, expression, cache)\tcp*{Generate the image}
            image $\rightarrow$
        out\_queue \tcp*{Return output image via queue}
        }
        \Else{
            mesh $\leftarrow$ runFaceModel(viewpoint, expression)\tcp*{Face mesh from expr. parameters}
            cache $\leftarrow$ runDecoder(viewpoint, expression, mesh)\tcp*{Cache neural representation}
            "caching done" $\rightarrow$
        cache\_queue \tcp*{Notify cache completion via queue}
        }
    }
    \caption{Parallel Implementation (2x GPUs, 2x warp)}
    \label{algo:parallelAlgorithm}
\end{algorithm*}

\paragraph{Algorithm Time Complexity} Using the code provided by \href{https://github.com/Lyken17/pytorch-OpCounter}{pytorch-OpCounter}, we measured the FLOPs for both the generator and warping network which are 364G and 87G FLOPs respectively. This factor of 4x corresponds well with the 3.2x latency improvement that we reported. Note that our approach decouples the complexity of the generator and warp networks, i.e., the warp keeps the same time complexity when coupled with deeper generators (c.f.~experiment with ResNet in ablation study).

\section{Limitations of Capture Algorithm}
\begin{figure}[t!]
\begin{center}
\includegraphics[width=1\linewidth, trim={0cm 21.5cm 0cm 0cm},clip]{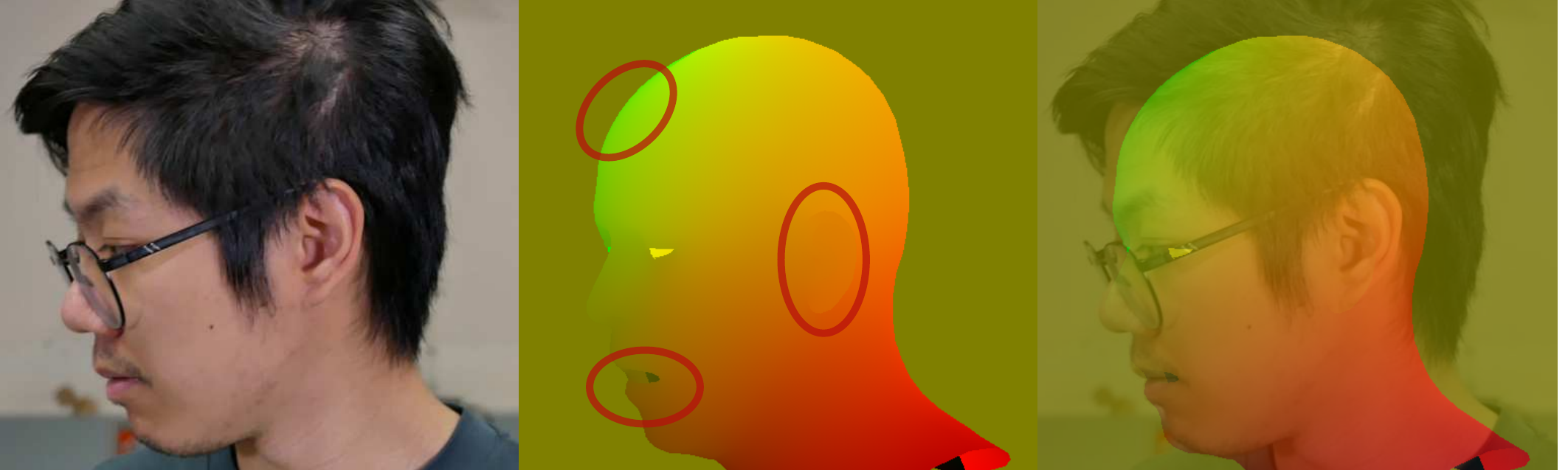}
\end{center}
\caption{\label{fig:limitation}
		{\bf Limitations of Head Reconstruction} We can see that at extreme angles and glasses present a problem to the capture algorithm and produces inaccurate UV maps and parameters.
	}
\end{figure}

Due to limitations with the DECA approach~\cite{DECA:Siggraph2021}, extreme head poses and cases when the subject is wearing glasses are not accurately reconstructed. Figure~\ref{fig:limitation} shows such a case with glasses and a perpendicular angle to the camera, leading to a misalignment of model and image around the mouth, ears, and forehead. 

\section{Additional Ablation on Caching Layers}

In this additional ablation, we sequentially remove the caching layers (C3, C4, C5) and measure the reconstruction quality in terms of L1 loss, PSNR, and SSIM. Each configuration was trained on the \textit{beard} dataset for 50 epochs, and metrics were computed on the held-out test set. Table~\ref{tab:ablation-cs} shows that adding the C4 cache layer is the most impactful, but since performance increases after adding all layers, we include all cache layers in our final configuration.

\begin{table}[]
\begin{center}
\begin{tabular}{ccc|ccc}
C3 & C4 & C5 & L1 $\downarrow$     & PSNR $\uparrow$   & SSIM $\uparrow$         \\ \hline \hline
$\checkmark$  &    &    & 0.0261 & 26.13 & 0.9048 \\
$\checkmark$  & $\checkmark$  &    & 0.0247 & 26.48 & 0.9096 \\
$\checkmark$  & $\checkmark$  & $\checkmark$  & \textbf{0.0246} & \textbf{26.54} & \textbf{0.9102} \\ \hline
\end{tabular}
\caption{\label{tab:ablation-cs}
		{\bf Ablating caching layers} for our warping network using the sequential variant (1x warp), showing the improvement after adding each component on the \emph{beard} dataset.} 
\end{center}
\end{table}

\section{Additional Ablation on the Warp Distance}

\begin{figure}
\begin{center}
\includegraphics[width=0.9\linewidth, trim={0cm 0cm 0cm 0cm},clip]{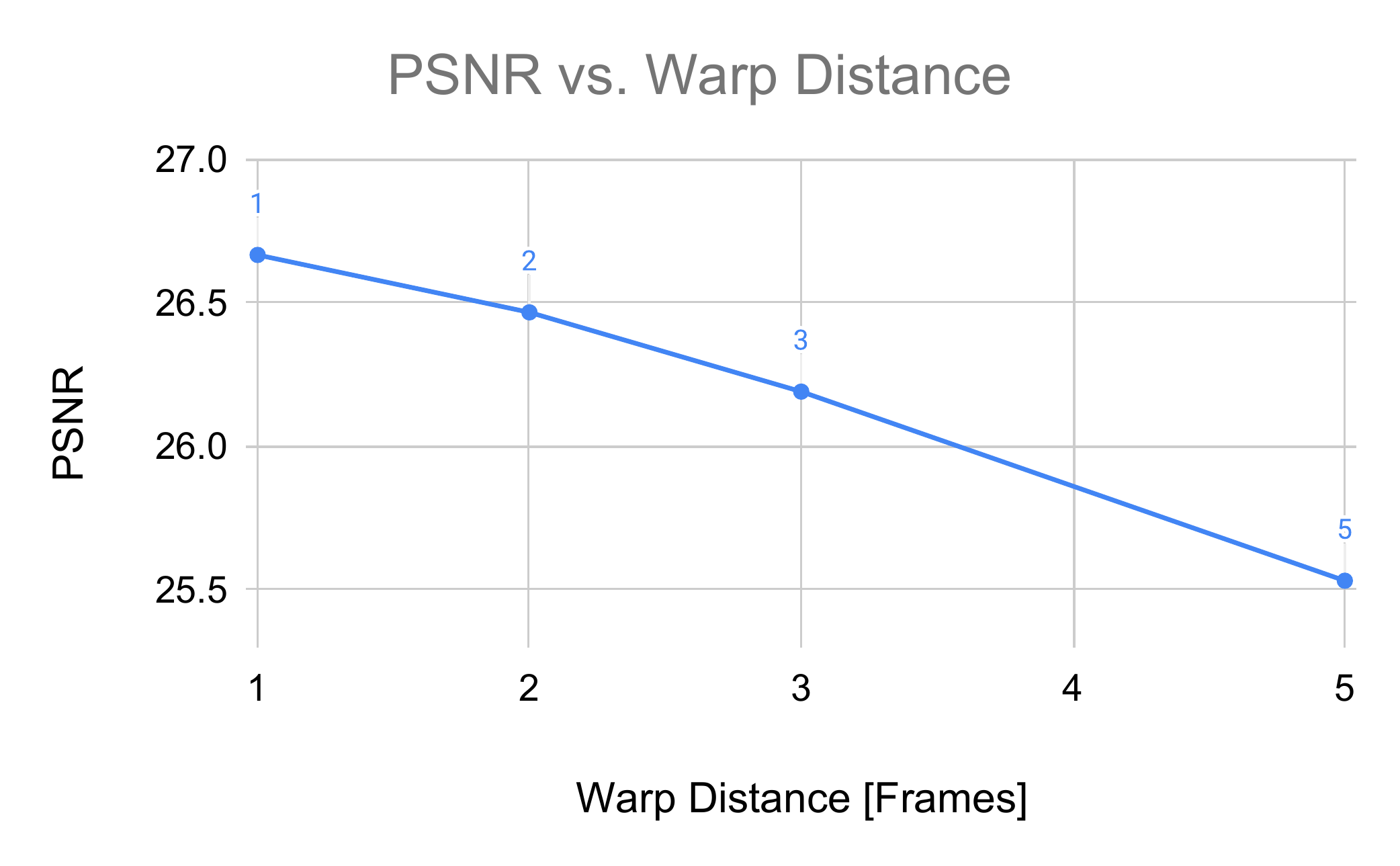}
\end{center}
\caption{\label{fig:warp_distance}
		{\bf Ablation on warp distance} as we increase the warp distance, or the number of warping operations per caching operation, the reconstruction quality decreases linearly.  
	}
\end{figure}

In the experiments in the main document, we present the results of our warping network by warping either 1 or 2 frames ahead. This ablation study measures the performance drop from warping more frames ahead.  Figure~\ref{fig:warp_distance} shows that the reconstruction quality decreases linearly with the warp distance on the \emph{beard} dataset, which was recorded at 30fps. Please note that this increase of error is expected as the proposed neural caching and warping is a local approximation designed for high-fps videos where the warp distance is small.

%% file: arxiv.bbl
\begin{thebibliography}{10}\itemsep=-1pt

\bibitem{blanz1999morphable}
Volker Blanz and Thomas Vetter.
\newblock A morphable model for the synthesis of {3D} faces.
\newblock 1999.

\bibitem{bulat2017far}
Adrian Bulat and Georgios Tzimiropoulos.
\newblock How far are we from solving the 2d \& 3d face alignment problem? (and
  a dataset of 230,000 3d facial landmarks).
\newblock In {\em International Conference on Computer Vision}, 2017.

\bibitem{carreira2018massively}
Joao Carreira, Viorica Patraucean, Laurent Mazare, Andrew Zisserman, and Simon
  Osindero.
\newblock Massively parallel video networks.
\newblock In {\em Proceedings of the European Conference on Computer Vision
  (ECCV)}, pages 649--666, 2018.

\bibitem{chu2021fast}
Xiangxiang Chu, Bo Zhang, Hailong Ma, Ruijun Xu, and Qingyuan Li.
\newblock Fast, accurate and lightweight super-resolution with neural
  architecture search.
\newblock In {\em 2020 25th International Conference on Pattern Recognition
  (ICPR)}, pages 59--64. IEEE, 2021.

\bibitem{deng2020disentangled}
Yu Deng, Jiaolong Yang, Dong Chen, Fang Wen, and Xin Tong.
\newblock Disentangled and controllable face image generation via {3D}
  imitative-contrastive learning.
\newblock In {\em CVPR}, 2020.

\bibitem{feichtenhofer2020x3d}
Christoph Feichtenhofer.
\newblock X3d: Expanding architectures for efficient video recognition.
\newblock In {\em Proceedings of the IEEE/CVF Conference on Computer Vision and
  Pattern Recognition}, pages 203--213, 2020.

\bibitem{DECA:Siggraph2021}
Yao Feng, Haiwen Feng, Michael~J. Black, and Timo Bolkart.
\newblock Learning an animatable detailed {3D} face model from in-the-wild
  images.
\newblock volume~40, 2021.

\bibitem{fried2019text}
Ohad Fried, Ayush Tewari, Michael Zollh{\"o}fer, Adam Finkelstein, Eli
  Shechtman, Dan~B Goldman, Kyle Genova, Zeyu Jin, Christian Theobalt, and
  Maneesh Agrawala.
\newblock Text-based editing of talking-head video.
\newblock {\em ACM TOG}, 2019.

\bibitem{geng2018warp}
Jiahao Geng, Tianjia Shao, Youyi Zheng, Yanlin Weng, and Kun Zhou.
\newblock Warp-guided {GANs} for single-photo facial animation.
\newblock {\em ACM TOG}, 2018.

\bibitem{ghosh2020gif}
Partha Ghosh, Pravir~Singh Gupta, Roy Uziel, Anurag Ranjan, Michael~J Black,
  and Timo Bolkart.
\newblock Gif: Generative interpretable faces.
\newblock In {\em 2020 International Conference on 3D Vision (3DV)}, pages
  868--878. IEEE, 2020.

\bibitem{hu2020temporally}
Ping Hu, Fabian Caba, Oliver Wang, Zhe Lin, Stan Sclaroff, and Federico
  Perazzi.
\newblock Temporally distributed networks for fast video semantic segmentation.
\newblock In {\em Proceedings of the IEEE/CVF Conference on Computer Vision and
  Pattern Recognition}, pages 8818--8827, 2020.

\bibitem{pix2pix2017}
Phillip Isola, Jun-Yan Zhu, Tinghui Zhou, and Alexei~A Efros.
\newblock Image-to-image translation with conditional adversarial networks.
\newblock 2017.

\bibitem{johnson2016perceptual}
Justin Johnson, Alexandre Alahi, and Li Fei-Fei.
\newblock Perceptual losses for real-time style transfer and super-resolution.
\newblock In {\em European conference on computer vision}, pages 694--711.
  Springer, 2016.

\bibitem{karras2021alias}
Tero Karras, Miika Aittala, Samuli Laine, Erik H{\"a}rk{\"o}nen, Janne
  Hellsten, Jaakko Lehtinen, and Timo Aila.
\newblock Alias-free generative adversarial networks.
\newblock {\em arXiv preprint arXiv:2106.12423}, 2021.

\bibitem{karras2018style}
Tero Karras, Samuli Laine, and Timo Aila.
\newblock A style-based generator architecture for generative adversarial
  networks.
\newblock In {\em CVPR}, 2019.

\bibitem{karras2020analyzing}
Tero Karras, Samuli Laine, Miika Aittala, Janne Hellsten, Jaakko Lehtinen, and
  Timo Aila.
\newblock Analyzing and improving the image quality of {StyleGAN}.
\newblock In {\em CVPR}, 2020.

\bibitem{kingma2014adam}
Diederik Kingma and Jimmy Ba.
\newblock Adam: A method for stochastic optimization.
\newblock In {\em ICLR}, 2015.

\bibitem{li2017learning}
Tianye Li, Timo Bolkart, Michael~J Black, Hao Li, and Javier Romero.
\newblock Learning a model of facial shape and expression from 4d scans.
\newblock {\em ACM Trans. Graph.}, 36(6):194--1, 2017.

\bibitem{malinowski2020sideways}
Mateusz Malinowski, Grzegorz Swirszcz, Joao Carreira, and Viorica Patraucean.
\newblock Sideways: Depth-parallel training of video models.
\newblock In {\em Proceedings of the IEEE/CVF Conference on Computer Vision and
  Pattern Recognition}, pages 11834--11843, 2020.

\bibitem{mildenhall2020nerf}
Ben Mildenhall, Pratul~P Srinivasan, Matthew Tancik, Jonathan~T Barron, Ravi
  Ramamoorthi, and Ren Ng.
\newblock Nerf: Representing scenes as neural radiance fields for view
  synthesis.
\newblock In {\em European conference on computer vision}, pages 405--421.
  Springer, 2020.

\bibitem{nagano2018pagan}
Koki Nagano, Jaewoo Seo, Jun Xing, Lingyu Wei, Zimo Li, Shunsuke Saito, Aviral
  Agarwal, Jens Fursund, and Hao Li.
\newblock {paGAN}: real-time avatars using dynamic textures.
\newblock {\em ACM TOG}, 2018.

\bibitem{olszewski2017realistic}
Kyle Olszewski, Zimo Li, Chao Yang, Yi Zhou, Ronald Yu, Zeng Huang, Sitao
  Xiang, Shunsuke Saito, Pushmeet Kohli, and Hao Li.
\newblock Realistic dynamic facial textures from a single image using {GANs}.
\newblock In {\em ICCV}, 2017.

\bibitem{suwajanakorn2017synthesizing}
Supasorn Suwajanakorn, Steven~M Seitz, and Ira Kemelmacher-Shlizerman.
\newblock Synthesizing {Obama}: learning lip sync from audio.
\newblock {\em ACM TOG}, 2017.

\bibitem{tewari2020stylerig}
Ayush Tewari, Mohamed Elgharib, Gaurav Bharaj, Florian Bernard, Hans-Peter
  Seidel, Patrick P{\'e}rez, Michael Z{\"o}llhofer, and Christian Theobalt.
\newblock {StyleRig}: Rigging {StyleGAN} for 3d control over portrait images.
\newblock In {\em CVPR}, 2020.

\bibitem{thies2019deferred}
Justus Thies, Michael Zollh{\"o}fer, and Matthias Nie{\ss}ner.
\newblock Deferred neural rendering: Image synthesis using neural textures.
\newblock {\em ACM Transactions on Graphics (TOG)}, 38(4):1--12, 2019.

\bibitem{thies2015real}
Justus Thies, Michael Zollh{\"o}fer, Matthias Nie{\ss}ner, Levi Valgaerts, Marc
  Stamminger, and Christian Theobalt.
\newblock Real-time expression transfer for facial reenactment.
\newblock {\em ACM TOG}, 2015.

\bibitem{thies2016face2face}
Justus Thies, Michael Zollhofer, Marc Stamminger, Christian Theobalt, and
  Matthias Nie{\ss}ner.
\newblock {Face2Face}: Real-time face capture and reenactment of rgb videos.
\newblock In {\em CVPR}, 2016.

\bibitem{van2016asynchronous}
JMP Van~Waveren.
\newblock The asynchronous time warp for virtual reality on consumer hardware.
\newblock In {\em Proceedings of the 22nd ACM Conference on Virtual Reality
  Software and Technology}, pages 37--46, 2016.

\bibitem{vaswani2017attention}
Ashish Vaswani, Noam Shazeer, Niki Parmar, Jakob Uszkoreit, Llion Jones,
  Aidan~N Gomez, {\L}ukasz Kaiser, and Illia Polosukhin.
\newblock Attention is all you need.
\newblock In {\em NeurIPS}, 2017.

\bibitem{vlasic2005face}
Daniel Vlasic, Matthew Brand, Hanspeter Pfister, and Jovan Popovic.
\newblock Face transfer with multilinear models.
\newblock {\em ACM TOG}, 2005.

\bibitem{wang2021one}
Ting-Chun Wang, Arun Mallya, and Ming-Yu Liu.
\newblock One-shot free-view neural talking-head synthesis for video
  conferencing.
\newblock In {\em Proceedings of the IEEE/CVF Conference on Computer Vision and
  Pattern Recognition}, pages 10039--10049, 2021.

\bibitem{yu2021plenoctrees}
Alex Yu, Ruilong Li, Matthew Tancik, Hao Li, Ren Ng, and Angjoo Kanazawa.
\newblock Plenoctrees for real-time rendering of neural radiance fields.
\newblock {\em arXiv preprint arXiv:2103.14024}, 2021.

\bibitem{zakharov2020fast}
Egor Zakharov, Aleksei Ivakhnenko, Aliaksandra Shysheya, and Victor Lempitsky.
\newblock Fast bi-layer neural synthesis of one-shot realistic head avatars.
\newblock In {\em ECCV}, 2020.

\bibitem{zhou2019evaluation}
Qian Zhou, Georg Hagemann, Dylan Fafard, Ian Stavness, and Sidney Fels.
\newblock An evaluation of depth and size perception on a spherical fish tank
  virtual reality display.
\newblock {\em IEEE transactions on visualization and computer graphics}, 2019.

\bibitem{zhou2017error}
Qian Zhou, Gregor Miller, Kai Wu, Ian Stavness, and Sidney Fels.
\newblock Analysis and practical minimization of registration error in a
  spherical fish tank virtual reality system.
\newblock volume 10114, pages 519--534, 03 2017.

\bibitem{zhou20173dps}
Qian Zhou, Kai Wu, Gregor Miller, Ian Stavness, and Sidney Fels.
\newblock 3dps: An auto-calibrated three-dimensional perspective-corrected
  spherical display.
\newblock In {\em 2017 IEEE Virtual Reality (VR)}, pages 455--456. IEEE, 2017.

\end{thebibliography}
